\documentclass{article}
\usepackage[T1]{fontenc}
\usepackage{iclr2027_conference,times}
\iclrfinalcopy
\usepackage{amsmath,amssymb,amsthm,booktabs,graphicx,etoolbox,seqsplit,array,placeins,float}
\usepackage{xurl}
\usepackage[hidelinks]{hyperref}
\usepackage{url}
\hypersetup{pdftitle={Shared Global KV with Layer-Specific Local History},
 pdfauthor={Xinglang Xian},pdfsubject={Research preprint}}
\newtheorem{proposition}{Proposition}

\newcommand{\CLABaseline}{GQA4--CLA2}

\newcommand{\RMS}{\operatorname{RMSNorm}}

\newcommand{\KVSharedMiB}{33.50}
\newcommand{\KVIndependentMiB}{56.00}
\newcommand{\KVReductionFactor}{1.67}
\newcommand{\KVAsymptoticFactor}{1.78}

\newcommand{\ScaleTrainHours}{86.812735}
\newcommand{\ScaleExternalHours}{0.067523}

\newcommand{\NormDevHistoryPpl}{1.37}

\newcommand{\NormBookHistoryPpl}{1.08}
\newcommand{\NormGPUHours}{171.48}

\newcommand{\HoldoutTargets}{2,095,344}
\newcommand{\HoldoutDocuments}{1,809}
\newcommand{\BudgetWoneDevPpl}{1.11}
\newcommand{\BudgetCountDevPpl}{0.95}

\newcommand{\BudgetComputeBookPpl}{1.76}

\newcommand{\BudgetComputeTestPpl}{1.02}

\newcommand{\TestJointPpl}{1.41}

\newcommand{\SourceInputDevPpl}{0.03}

\newcommand{\SourceInputDevDelta}{0.000256}
\newcommand{\SourceKVDevPpl}{0.37}

\newcommand{\SourceKVDevDelta}{0.003649}

\newcommand{\HistoryDevIndependentGain}{0.009530}
\newcommand{\HistoryDevPairedGain}{0.008722}

\newcommand{\HistoryDevIndependentPpl}{0.95}
\newcommand{\HistoryDevPairedPpl}{0.87}

\newcommand{\UnifiedTwoRequestMin}{1.088}
\newcommand{\UnifiedTwoRequestMax}{1.180}

\newcommand{\UnifiedFourRequestMin}{1.082}
\newcommand{\UnifiedFourRequestMax}{1.184}

\newcommand{\ClosureInputPrefillRatio}{0.857}
\newcommand{\ClosureInputPrefillSaving}{14.3}
\newcommand{\ClosureInputRequestRatio}{0.999}

\newcommand{\ClosureInputRequestMin}{0.974}
\newcommand{\ClosureInputRequestMax}{1.016}
\newcommand{\ClosureKVPrefillRatio}{0.853}

\newcommand{\ClosureKVRequestRatio}{0.890}
\newcommand{\ClosureKVRequestSaving}{11.0}

\newcommand{\ClosureKVCacheSaving}{3.0}
\newcommand{\ClosureKVRequestMin}{0.867}
\newcommand{\ClosureKVRequestMax}{0.895}

\newcommand{\FinalConfirmGQA}{1.43}
\newcommand{\FinalConfirmCLA}{1.15}
\newcommand{\FinalBooksGQA}{6.96}
\newcommand{\FinalBooksCLA}{3.97}
\newcommand{\FinalCachePremium}{19.63}
\newcommand{\FinalPrefillSaving}{16.40}
\newcommand{\FinalRequestPremium}{32.98}

\newcommand{\ContextLongGain}{1.16}
\newcommand{\ContextBookGain}{2.91}
\newcommand{\ContextShortGain}{1.14}

\title{Shared Global KV\\with Layer-Specific Local History}
\author{Xinglang Xian}
\begin{document}
\raggedbottom
\maketitle

\begin{abstract}
Decoder-only Transformer language models cache keys and values (KV) to reuse
past computation during generation. Sharing KV across layers saves storage but
reduces the diversity of representations available across depth. We study what
local memory should retain alongside shared global KV, separating historical
content from the input source used to form it.
At 126M parameters and 2K context, an eight-seed study finds about 1.4\%
lower held-out test perplexity with local history than with a current-token
local branch. Capacity, entry-count and training-compute controls support the
value of historical content. In a two-seed comparison, this value persists
when adjacent layers share local inputs while retaining independent projections;
source sharing also shortens exact cache-construction dependencies.
Against GQA and adjacent-layer KV sharing, equal bounded learning-rate searches
and new-seed confirmation yield better same-source likelihood with larger
caches and higher long-request latency. The ordering against adjacent-layer
sharing persists after equal-token adaptation to 8K, with a short-context cost.
The eight-seed external-book history effect remains uncertain, and downstream
outcomes vary by task. We derive a sufficient suffix schedule that reduces
upper-layer construction work while preserving the complete cache in exact
arithmetic.
\end{abstract}

\section{Introduction}
\label{sec:intro}
Decoder-only Transformer language models generate tokens autoregressively.
They cache the keys and values (KV) of preceding tokens so that each new token
can attend to the past without recomputing those representations. Keeping a
separate cache at every layer makes storage grow with both sequence length
and model depth. Cross-layer sharing reduces this storage by reusing KV
representations \citep{yoco,cla,kvstudy}. It preserves access to past positions,
but gives layers fewer distinct representations of those positions to read.
This raises a design question: \emph{which recent representations are worth
retaining alongside a shared global memory?}

\begin{figure}[t]
\centering
\includegraphics[width=\linewidth]{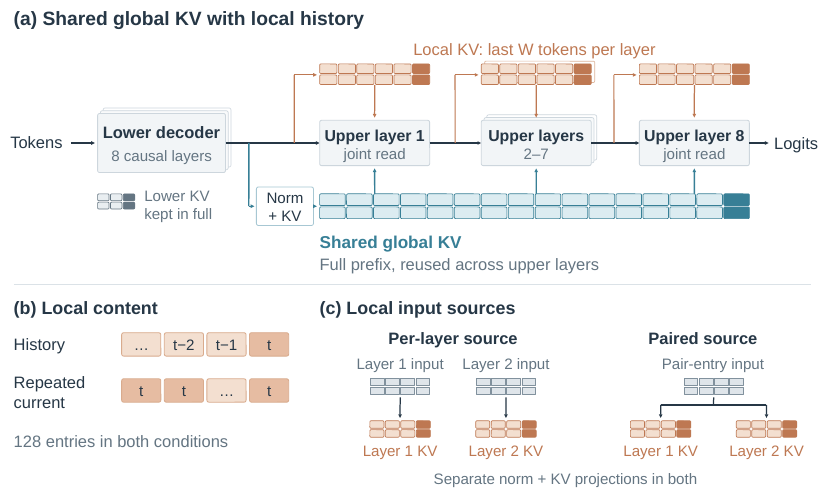}
\caption{\textbf{Content retention and local input sources are separate design choices.}
(a) Eight lower layers supply a normalized global KV projection reused by eight
upper layers. Each upper layer reads this bank jointly with its own local KV
through one softmax. The middle group abbreviates layers 2--7; lower KV is
retained in full. (b) History and repeated-current entries have the same count,
shown for $W=128$. (c) Two adjacent upper layers take either separate inputs or
the same pair-entry representation. Each arrow applies an independent
normalization and KV projection; queries, residuals and FFNs remain
layer-specific. Blue/orange denote global/local KV, and gray matrices denote
input representations. Cells are schematic.}
\label{fig:architecture}
\end{figure}

We study this question in a non-looped decoder with shared global KV and
bounded, layer-specific local KV (Figure~\ref{fig:architecture}). A completed
lower causal stack supplies the global bank. Each upper layer forms its own
query and local projections, then reads both banks through one softmax.
This allocation provides full causal coverage and a short history of upper-stack
features. We then vary their content and input source separately: retaining
past entries and refreshing their source at every layer are different choices.

Shared global KV and private local history also coexist in Parallel Loop
Transformer (PLT) \citep{plt}. We investigate what this allocation should
retain by separating \emph{historical content} from \emph{input-source granularity}.
In the tested adjacent-pair setting, history remains useful when receivers
share local inputs but keep independent projections. This reuse shortens
construction dependencies without reducing the number of saved local banks:
sharing the source of a representation and sharing the representation itself
have different consequences.

The same allocation has an execution consequence. Bounded upper dependencies
allow complete-cache construction from progressively shorter suffixes.
We therefore connect predictive comparisons to the storage and request costs
of the same checkpoints, and distinguish the cost of adding history from the
saving obtained by executing its construction more efficiently.

\newpage
\paragraph{1. Historical content and representation source.}
An eight-seed factorial finds about 1.4\% lower reserved-test perplexity with
joint history, supported by capacity, entry-count and training-compute controls.
A two-seed source comparison retains a history benefit under adjacent-pair
inputs with separate projections, connecting input reuse to shorter construction
dependencies (Section~\ref{sec:experiments}).

\paragraph{2. Quality, state and construction cost.}
After equal three-point learning-rate searches, the design improves same-source
likelihood over GQA2 and GQA4--CLA2 in both new seeds. The CLA ordering persists
after equal-token 8K adaptation. Complete-cache and request measurements quantify
the resource cost; a sufficient suffix schedule reduces construction work while
preserving the full cache (Sections~\ref{sec:gqa} and~\ref{sec:exact}).

\section{Related work}
\label{sec:related}
\paragraph{Source and placement of shared KV.}
Multi-query attention (MQA) and grouped-query attention (GQA) share KV across
heads \citep{mqa,gqa}. YOCO splits the decoder into lower and upper stacks,
sharing global KV with upper-layer prefill early exit \citep{yoco}; LCKV uses
upper-layer KV and iterative training \citep{lckv}. The pizza-bottom design
also provides a shared global bank \citep{kvstudy}. CLA studies sharing factors
and learning rates \citep{cla}, while FusedKV reconstructs upper KV from lower
caches \citep{fusedkv}. We separate local content from input-source reuse
alongside a shared global bank.

\paragraph{Local information and branch fusion.}
Longformer jointly normalizes global-token and window scores \citep{longformer}.
Block Transformer combines global blocks with local decoding \citep{blocktransformer};
MixAttention combines windows, full attention and KV reuse \citep{mixattention}.
Hymba couples attention and state-space heads with global/local attention and
cross-layer sharing \citep{hymba}. PLT combines shared first-loop KV with private
sliding windows through a query-dependent head gate \citep{plt}. It targets
weight-tied loops; we study content and source reuse in an untied stack. Our
separate-fusion control uses query-independent weights and does not test PLT's
dynamic gate. CLSA shares sparse selection \citep{clsa}; DeepSeek-V4.1-Flash
uses global/local separation and bounded replay \citep{deepseek}. Gated
Attention learns gates on SDPA outputs \citep{gatedattention}; our joint
normalization instead determines branch weights from attention scores.

\paragraph{Execution and retention costs.}
FlashAttention reduces memory traffic \citep{flashattention}. GoldFinch combines
finite-tail execution with compressed global representations \citep{goldfinch}.
Our schedule separates KV-input and query/output suffixes to reduce construction
work while preserving the complete cache. UniPrefill selects token work;
POP separates prefill pruning from cache production \citep{uniprefill,pop}.
Attention Residuals changes depth aggregation \citep{attnres}; we study temporal
entries under shared global KV.

\section{Method: architecture and design interventions}
\label{sec:backbone}
A shared global bank provides full causal coverage; bounded local banks retain
recent features (Figure~\ref{fig:architecture}). Each upper layer has its own
local projection and bank, with either layer-specific or shared inputs.
We intervene on content, source, capacity, entry count and fusion.

The primary model has 126M parameters, 16 causal layers and 2K packed inputs.
It uses RMSNorm, SwiGLU and rotary positions \citep{rmsnorm,swiglu,rope}; full
dimensions appear in Table~\ref{tab:modelshapes}. Attention and loss are
document-isolated, and rotary positions restart at document boundaries.
Normalization, residual additions and feedforward layers are positionwise.

\paragraph{Shared and local representations.}
Eight lower layers compute $H^0$ using causal grouped-query attention \citep{gqa}.
A normalized projection of $H^0$ produces $(K^g,V^g)$, shared by all eight upper
layers. Upper layer $j$ forms its own query $q^j_t$ and, when present, its local
pair $(k^j_i,v^j_i)$ from its normalized input. Each query reads two sources:
the global causal prefix and the receiving layer's recent local entries.
Let $G_t$ contain global positions at or before $t$ in the same document;
let $L_t(W)$ contain local positions in that document with $t-W<i\leq t$.
The superscripts $g$ and $j$ distinguish the shared and layer-specific sources.
Their scores enter a single normalization:
\begin{equation}
 a^j_t =
 \frac{\sum_{i\in G_t}e^{s^g_{ti}}v^g_i+
       \sum_{i\in L_t(W)}e^{s^j_{ti}}v^j_i}
      {\sum_{i\in G_t}e^{s^g_{ti}}+
       \sum_{i\in L_t(W)}e^{s^j_{ti}}},
 \qquad s_{ti}=q^j_t\mathbin{\cdot}k_i/\sqrt{64}.
 \label{eq:joint}
\end{equation}
Rotary position transforms are included in $q$ and $k$. A global and local entry
at the same source position are distinct representations in the same softmax;
their weights compete across branches. For each head, let $A_g,A_l$ denote
the separately normalized branch outputs. Equation~\ref{eq:joint} equivalently gives
\begin{equation}
 a=(1-\alpha)A_g+\alpha A_l,\qquad
 \alpha=\sigma\!\left(\operatorname{LSE}(s_l)-\operatorname{LSE}(s_g)\right),
 \label{eq:branchmass}
\end{equation}
where $\operatorname{LSE}(s)=\log\sum_i e^{s_i}$ and $\sigma$ is the logistic
function. The identity shows how branch scores determine local attention mass
using the existing attention parameters.

\paragraph{Interventions on local memory.}
We vary local content and its input source separately, then compare the complete
allocation with alternative ways to share KV (Table~\ref{tab:controls}). Throughout, the
\emph{history model} uses a 128-position local window, including the current
token. The \emph{current-only control} retains just the current local pair;
its global branch still reads the complete causal prefix. We name controls
by their intervention; Appendix~\ref{app:controlnames} maps these names to the
original experiment IDs. Common tensors use seed-and-name keyed initialization.

\begin{table}[!ht]
\centering\small
\caption{\textbf{Comparisons separate content, source and resource allocation.}
Each row identifies a comparison and the evidence it provides. Training and
measurement protocols are matched within studies, as specified in the cited sections.}
\label{tab:controls}
\begin{tabular}{@{}>{\raggedright\arraybackslash}p{0.18\linewidth}>{\raggedright\arraybackslash}p{0.45\linewidth}>{\raggedright\arraybackslash}p{0.30\linewidth}@{}}
\toprule
Design factor & Comparison & Evidence \\
\midrule
Local content & History vs. current only, under joint and separate fusion & History benefit (\S\ref{sec:normalizationresults}) \\
\addlinespace[3pt]
Entry count & History vs. repeated current entries & Content beyond multiplicity (\S\ref{sec:budget}) \\
\addlinespace[3pt]
Training budget & History vs. current only with extra updates & Matched matrix-product budget (\S\ref{sec:budget}) \\
\addlinespace[3pt]
Input source & Per-layer vs. paired-layer inputs, crossed with entry content & Source granularity (\S\ref{sec:localsource}) \\
\addlinespace[3pt]
KV allocation & Shared global plus local history vs. GQA and GQA4--CLA2 & Quality, complete cache and request cost (\S\ref{sec:gqa}--\ref{sec:matchedcost}) \\
\bottomrule
\end{tabular}
\end{table}

\paragraph{An entry-multiplicity control.}
Replacing current-only entries with history changes both historical local content and local entry count in
Equation~\ref{eq:joint}. Define $m_t=|L_t(128)|$. The repeated-current control uses $m_t$
identical copies of the \emph{current} local pair. Their exact-arithmetic
contribution is equivalent to a single current entry with score
\begin{equation}
 \widetilde s^j_{tt}=s^j_{tt}+\log m_t.
 \label{eq:multiplicity}
\end{equation}
Counts follow the original mask, including short document prefixes. At fixed
scores, the offset increases local/global log odds by $\log m_t$
(Equation~\ref{eq:branchmass}); the local output remains the current value.
This replaces historical features at fixed multiplicity, with attention mass
free to adapt during training. All three content variants share parameter
count and the paired recipe (Appendix~\ref{app:pending}).

\paragraph{A fusion intervention.}
The factorial study crosses $W\in\{1,128\}$ with joint fusion ($J_W$) and
separate fusion ($S_W$). The latter computes
$a^j_t=(1-\beta_{jh})A_g+\beta_{jh}A_l$ per query head $h$, where
$\beta_{jh}=\sigma(b_{jh})$ is a learned, query-independent scalar initialized
to $1/2$. It adds 96 parameters and uses the same optimizer group as the other
weights. This weight is constant across queries, whereas the mass in
Equation~\ref{eq:branchmass} depends on their scores. All four cells are
independently trained.
Appendix~\ref{app:normalization} details the parameter roles of this intervention.

\section{Experiments}
\label{sec:experiments}
We test local content, input sources and the complete allocation
(Table~\ref{tab:controls}), then connect prediction to resource costs
(Section~\ref{sec:exact}).

\begin{figure}[!ht]
\centering
\includegraphics[width=\linewidth]{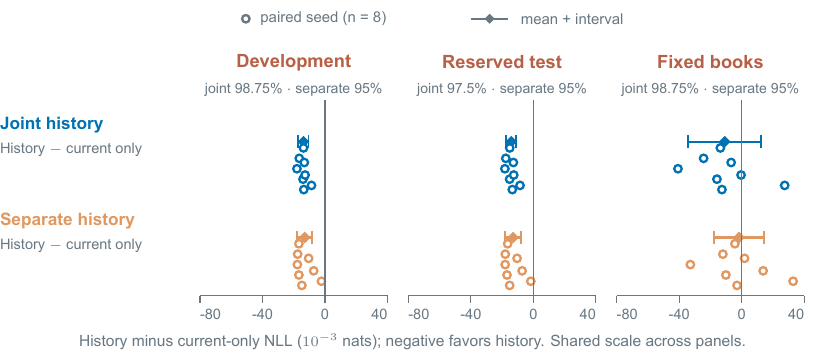}
\caption{\textbf{Local history improves development and reserved-test prediction
under both fusions.} Eight paired seeds use 126M models, 2K context and
1B training tokens. Open circles are seed effects; diamonds and bars are means
and paired-seed intervals at the levels printed above each panel. Primary
joint-history intervals are family-adjusted; separate-history intervals are
descriptive. Negative differences favor the history model. Fixed books show
greater seed variation. Fusion interactions are reported in
Appendix Figure~\ref{fig:normalization} and Table~\ref{tab:testcontrasts}.}
\label{fig:historyconfirmation}
\end{figure}

\subsection{Matched training and fixed evaluation}
The 2K studies train from random initialization on frozen FineWeb-Edu text
\citep{fineweb,finewebedu}, with the Mistral-7B-v0.3 tokenizer
\citep{mistraltokenizer}, 2K context and global batch 64. Pairing shares data order
and same-shaped named-tensor initialization; compute-matched arms can consume
different token totals. The factorial and
earlier controls use 1,000,079,360 tokens; the budget study uses 2.5B. Fixed
final checkpoints are independently reloaded for scoring.

Development scoring uses 2,095,344 valid targets. The external condition is
49 fixed PG-19 validation books \citep{compressive,pg19}, capped per book and
scored in reset 2K windows. A sampled training-overlap screen excludes one
book; remaining overlap is possible. Appendix~\ref{app:data} specifies the
caps, masks and screening. After freezing endpoints and comparisons, all
32 factorial and eight budget models also score reserved in-domain test text
(Appendix~\ref{app:holdout}). Training seeds are the independent units.
The factorial uses eight new seeds; its numerical recipe differs from earlier
controls as documented in Table~\ref{tab:recipes}.

\subsection{Local history across seeds and fusion recipes}
\label{sec:normalizationresults}
Local history improves development and reserved-test prediction under both
joint and learned separate fusion across all eight paired seeds
(Figure~\ref{fig:historyconfirmation}). Effects are history minus current-only NLL; $D_J$ denotes joint fusion.
Development and fixed-book primary contrasts use prespecified 98.75\%
paired-seed $t$ intervals, correcting for four comparisons
(Appendix~\ref{app:normresults}).

\paragraph{The in-domain benefit reproduces on reserved text.}
Joint history lowers development perplexity by \NormDevHistoryPpl\% and
reserved-test perplexity by \TestJointPpl\%. Every paired seed favors history
under both fusion recipes. The corresponding seed intervals exclude zero
(Figure~\ref{fig:historyconfirmation}); Tables~\ref{tab:normcontrasts}
and~\ref{tab:testcontrasts} give the complete contrasts. Perplexity reductions
use $100[1-\exp(\overline D_J)]$, averaging NLL differences over seeds before
conversion. The reserved-test primary comparisons use a separate 97.5\%
family-adjusted interval.

\paragraph{History gain and fusion quality answer different questions.}
With a 128-position window, joint fusion achieves lower NLL than learned
separate fusion on every seed in both development text and fixed books.
Yet the interaction---the difference between their history gains---remains
unresolved. Joint fusion predicts better in these runs; the experiment does
not establish that it increases the value of history.

\paragraph{External text reveals greater seed variation.}
Joint history improves on seven of eight seeds on fixed books, with a mean
perplexity reduction of \NormBookHistoryPpl\%, but the seed interval includes
zero. All leave-one-seed-out joint-history means remain negative: the mean
direction survives removal of any single seed, while uncertainty about the
external history effect remains. Appendix~\ref{app:normresults} reports the
interaction and sensitivity analyses alongside the seed effects.

\subsection{History under longer and matched computation budgets}
\label{sec:budget}
A useful local branch should retain its advantage when the current-token
alternative receives comparable training resources. We train models with history, current-only entries and
repeated-current entries for 2.5B tokens at 126M with two new paired seeds. A fourth arm gives the current-only control
additional updates to match the history model's counted training matrix-product FLOPs;
its schedule spans its own budget. This convention counts forward/backward
matrix products, with other operations and elapsed time accounted for
separately (Appendix~\ref{app:budgetprotocol}).

The history model improves over all three controls on both seeds in every scored condition
(Appendix Figure~\ref{fig:budget}). At the common token budget, development perplexity
falls by \BudgetWoneDevPpl\% against the current-only control and \BudgetCountDevPpl\% against
the repeated-current control, supporting historical content beyond entry multiplicity.
Against the compute-matched current-only control, the reductions are \BudgetComputeTestPpl\%
on reserved test and \BudgetComputeBookPpl\% on fixed books: the history model retains an advantage at matched counted computation.
Appendix~\ref{app:budgetprotocol} reports all endpoints and paired differences.

\paragraph{From likelihood to task predictions.}
Fixed zero-shot PIQA, HellaSwag and ARC-Easy evaluations
\citep{piqa,hellaswag,arcbenchmark} give mixed outcomes against the compute-matched current-only control:
$+0.54$, $-0.03$ and $-0.72$ percentage points. All descriptive
paired-example intervals include zero (Appendix~\ref{app:downstream}).

\subsection{How finely must local sources be separated?}
\label{sec:localsource}
\begin{figure}[!ht]
\centering
\includegraphics[width=\linewidth]{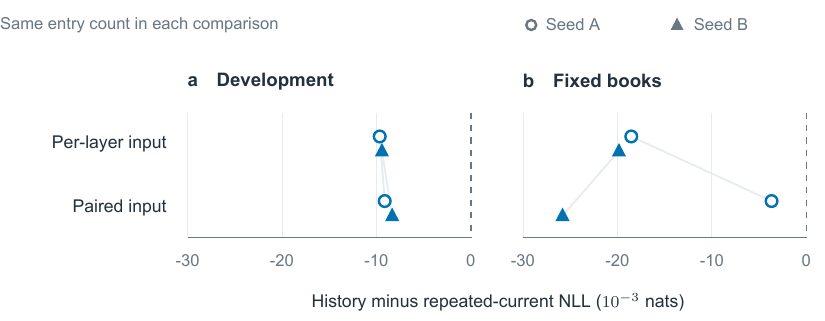}
\caption{\textbf{History remains useful when adjacent layers share its input source.}
Two paired seeds use 126M models, 2K context and 2.5B tokens.
Marks show history-minus-repeated-current NLL; negative values favor history.
Lines connect a seed across input sources (Figure~\ref{fig:architecture}(c)).
Both panels share a scale. The book responses differ across seeds; this
comparison does not establish source equivalence.}
\label{fig:localsource}
\end{figure}

History length and representation source are separate choices. At 2.5B tokens,
we cross per-layer versus adjacent-pair input sources with 128-position history versus
repeated-current entries, retaining receiver-specific projections.
Within each adjacent pair, both receivers form local KV from the pair-entry
representation, using separate RMSNorm and KV projections. Queries, residual
streams and feedforward blocks remain layer-specific. Repeated-current
controls entry count under each source.
History lowers development perplexity by \HistoryDevIndependentPpl\% with
per-layer inputs and \HistoryDevPairedPpl\% with paired inputs, averaging the
two seeds. Both seeds favor history under both sources on development text
and fixed books (Figure~\ref{fig:localsource}); the book interactions differ
in direction. These observations retain the value of earlier content under
paired inputs, while leaving source equivalence and interaction unresolved.

\paragraph{Sharing inputs and sharing banks have different consequences.}
For the history model, paired inputs change mean development perplexity by
+\SourceInputDevPpl\%. They shorten construction dependencies from layers to
groups while retaining one local bank per receiver. At the FP32 reference
workload, this reduces prefill time by \ClosureInputPrefillSaving\%, with little change in complete
request time. Sharing the projected KV bank instead raises development
perplexity by \SourceKVDevPpl\%, changes capacity and computation, and saves
\ClosureKVCacheSaving\% of complete cache and \ClosureKVRequestSaving\% of
request time (Appendix~\ref{app:sourcecost}). Thus input-source reuse changes
construction work; bank reuse additionally changes retained state.

We next compare the complete allocation with standard ways of sharing KV.
Capacity, a 305M early-training probe and an alternative global provider
appear in Appendices~\ref{app:historical}--\ref{app:providerextension}.

\subsection{The complete design against GQA and cross-layer sharing}
\label{sec:gqa}
We compare local history with all-layer GQA2 and adjacent-layer KV sharing
(\CLABaseline{}; \citealp{cla}): four KV heads shared by each pair of layers.
Its eight full-length banks match GQA2's logical cache size. All use matched
matrix-product training budgets. Each architecture receives the same three-point
learning-rate search, followed by training on two new seeds. All select the
upper grid boundary, so the results establish the ordering within this shared
search budget (Appendix~\ref{app:finallr}).

On the new same-source confirmation text (two seeds), local history lowers perplexity by
\FinalConfirmGQA\% against GQA2 and \FinalConfirmCLA\% against \CLABaseline{},
with both seeds favoring history. Fixed-book reductions are \FinalBooksGQA\%
and \FinalBooksCLA\%, also in both seeds. These whole-design results use different checkpoints and text from the
eight-seed history-versus-current reserved-test comparison. Downstream results vary:
PIQA favors the alternatives, HellaSwag favors history, and ARC-Easy changes
direction across seeds. Section~\ref{sec:matchedcost} measures the resources
of these same six checkpoints. Earlier common-recipe comparisons, including
GQA4, remain in Appendices~\ref{app:gqa} and~\ref{app:clacommon}.

\begin{figure}[!ht]
\centering
\includegraphics[width=\linewidth]{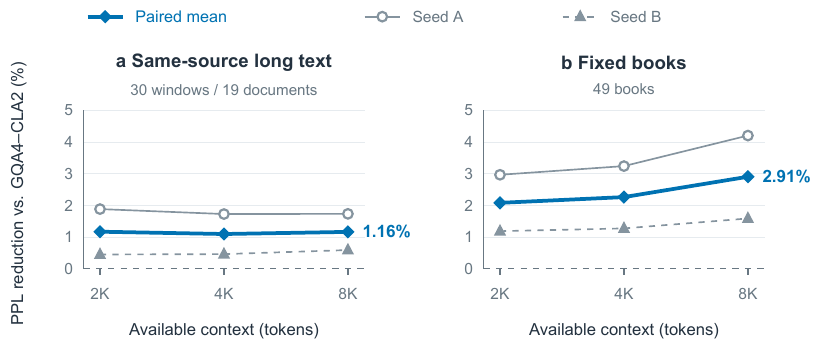}
\caption{\textbf{The architecture advantage persists to 8K and widens on fixed books.}
Four 126M checkpoints adapted on 134M tokens at 8K score identical final
1,024 targets at every length, with local window 128. Positive values favor
local history over \CLABaseline{}. Gray marks are paired seeds; diamonds
convert their mean NLL difference to PPL reduction. This whole-architecture
comparison and its short-context cost are detailed in Appendix~\ref{app:contextadaptation}.}
\label{fig:contextadaptation}
\end{figure}

\paragraph{Extending the comparison to 8K.}
The quality ordering also persists after equal-token context adaptation.
We continue both selected-rate history and \CLABaseline{} checkpoints,
using the same position interpolation \citep{positioninterpolation} and 134M
adaptation tokens per model. On identical targets at 2K, 4K and 8K, the mean
advantage stays near 1.1--1.2\% on same-source long text and grows from
2.08\% to \ContextBookGain\% on fixed books (Figure~\ref{fig:contextadaptation});
both paired seeds favor history at every length. On original 2K development
text, both incur about 2.8\% higher perplexity than their native parents,
while history retains its lead. Appendix~\ref{app:contextadaptation} reports
absolute scores and the adaptation protocol.

\section{Cache construction and inference cost}
\label{sec:exact}
Architecture comparisons use each model's exact construction route; execution
comparisons hold the model fixed.

\subsection{Architecture trade-offs at the same checkpoints}
\label{sec:matchedcost}
\begin{figure}[!ht]
\centering
\includegraphics[width=\linewidth]{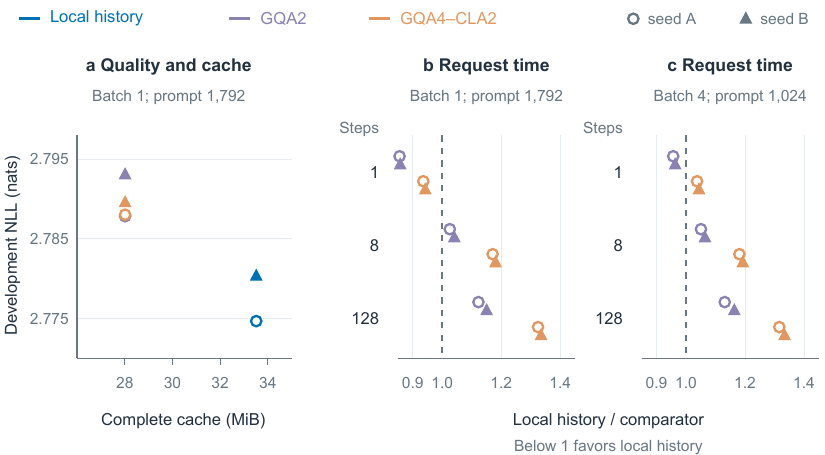}
\caption{\textbf{The history model improves likelihood with more cache; request cost depends on the workload.}
Selected-rate 126M models use matched training matrix-product budgets.
(a) Development NLL versus complete cache, including document IDs; lower is better.
(b--c) Request-time ratios below one favor local history. Circles/triangles
identify confirmation seeds; ratios aggregate three timing blocks without
interpolation. Execution uses FP32 IEEE, graph prefill and eager teacher-forced decode.}
\label{fig:gqatradeoff}
\end{figure}

\begin{samepage}
Figure~\ref{fig:gqatradeoff} measures all six selected-rate confirmation
checkpoints with complete caches, FP32 IEEE, graph-accelerated exact prefill
and eager decode, rescoring development NLL in this precision. The baseline
also skips unnecessary final-pair query/output work. Appendix~\ref{app:finallr}
details numerical qualification and three rotated timing blocks.

At batch one and 1,792 prompt tokens, history uses \FinalCachePremium\% more
cache than either alternative: 3.5 MiB for one additional full-length bank
and 2 MiB for local windows. Relative to \CLABaseline{}, prefill takes
\FinalPrefillSaving\% less time, but a request with 128 further decode steps
takes \FinalRequestPremium\% more. History is faster with one extra step;
CLA is faster with eight, and already with one at batch four and 1,024 prompt
tokens. Appendix~\ref{app:finallr} reports all six workloads and absolute times
(Figure~\ref{fig:requestduration}); the separate local-source/GQA4 campaign
is in Appendix~\ref{app:unifiedcost}.
\end{samepage}

\paragraph{Separating the local-history decision.}
Within the shared backbone, adding history improves FP32 perplexity by
1.11\%, with 31.1\% more prefill time and 6.3\% more cache at the same workload.
In this separate two-seed, 2.5B-token study, exact construction gives 1.82 times the prefill speed
of full upper execution but 1.014 times the request speed
(Appendix~\ref{app:matchedcost}, Figure~\ref{fig:qualitycost}).

\subsection{Why bounded history permits exact suffix execution}
With lower/global state complete, upper dependencies propagate through bounded
local windows. KV projection precedes attention output, allowing different
KV-input and query/output suffixes. GoldFinch uses a final $2G-1$-token region
for $G$ GOLD layers with token shifts \citep{goldfinch}; our schedule accounts
for the local dependencies and complete saved cache in this decoder.

\begin{samepage}
\begin{proposition}[Sufficient full-cache prefill schedule]
\label{prop:tail}
Consider $U$ upper layers with an inclusive local window $W$, fixed complete
global KV, causal document-isolated attention, and positionwise normalization,
residual and feedforward operations. For a prompt of length $N$, at upper layer
$j\in\{1,\ldots,U\}$ retain suffix lengths
\begin{align}
 r_{\rm in}(j)&=\min\{N,1+(U-j+1)(W-1)\},\label{eq:rin}\\
 r_{\rm out}(j)&=\min\{N,1+(U-j)(W-1)\}.\label{eq:rout}
\end{align}
Project local KV from the last $r_{\rm in}(j)$ inputs and save their last
$\min(N,W)$ entries. Compute queries, attention outputs, residuals and feedforward
outputs only for the last $r_{\rm out}(j)$ positions. With original positions
and masks, this returns the same prompt-end output and complete saved cache
as full upper prefill in exact arithmetic.
\end{proposition}
\end{samepage}

\begin{proof}[Proof sketch]
Work backward from the final token. With global KV fixed, $r$ output positions
need at most $\min(N,r+W-1)$ predecessor inputs, giving
Equations~\ref{eq:rin}--\ref{eq:rout}. Every KV-input suffix covers the final
$\min(N,W)$ saved entries. Positionwise operations add no dependencies, and
document masks only remove edges. Forward induction therefore preserves the
retained representations and their cached projections. The full argument is
in Appendix~\ref{app:proof}.
\end{proof}

For $U=8$, $W=128$ and $N\geq1017$, the schedule uses 4,580 KV-input
and 3,564 query/output positions, versus 8,136 of each for a uniform sufficient
suffix of 1,017. Appendix Figure~\ref{fig:schedule} shows every layer's suffix.
All exact routes retain the same complete KV; timings include the shared lower
computation. The global-only control and the current-only control use their stronger last-position upper execution paths.

\subsection{Exact prefill with equally accelerated controls}
\label{sec:prefillresults}
Two earlier 126M checkpoints compare layerwise, full and uniform-suffix
prefill at three lengths with rotated repeats. All six routes use CUDA Graphs
and include lower/global computation, complete cache and prompt-end head
(Appendix~\ref{app:prefillqualification}).

At 1,792 tokens, layerwise exact prefill is about 1.81 times as fast as full
prefill and 1.31 times as fast as the uniform sufficient suffix
(Table~\ref{tab:prefill}), holding the history model fixed.
History still costs approximately 31\% more prefill time than current-only
and 33\% more than global-only (Appendix Figure~\ref{fig:graph}).

\subsection{Complete storage and paused-state retention}
\label{sec:storage}
Each lower layer and the shared provider retain full-length KV; only the
upper local banks have bounded storage. With eight lower and eight upper
layers, the reduction relative to independent full KV approaches $16/9$
as context grows. Appendix~\ref{app:retention} gives the complete formula,
finite-length counts and document metadata; weights and workspaces are separate.

\paragraph{How to retain the local state.}
Local CPU offload uses less paused GPU memory and restores faster than
boundary replay. Approximate 128-position replay closely recovers predictions,
but retains 0.375 MiB more GPU state and takes 1.49--1.50 times as long to
restore under immediate resume (Appendix~\ref{app:retention}).

\section{Discussion}
\label{sec:limitations}
\paragraph{Retaining content and choosing its source.}
The tested paired-input design preserves a history benefit and shortens
construction dependencies while retaining one local KV bank per receiver.
Sharing those banks instead changes both prediction and storage. Input-source
reuse and KV reuse therefore require different resource comparisons.
Joint fusion gives lower absolute loss, while its interaction with the
history benefit remains unresolved.

\paragraph{Choosing an allocation for a workload.}
The measured choice is between lower language-model loss and lower resource
cost. When prefill dominates, the history model's shorter construction can
reduce request time; at the tested batch-one shape, the ordering already
reverses between one and eight extra steps. Longer outputs favor \CLABaseline{}.
For input-source sharing within the history model, prefill savings do not imply
smaller caches. These are workload-specific choices under the measured FP32
implementation; downstream scores do not yet establish an application benefit.

\paragraph{Scope and next comparisons.}
The core content study uses 126M models at 2K; the two-seed architecture
comparison extends to 8K after adaptation. The eight-seed within-backbone
book interval includes zero; the two selected-rate architecture seeds favor
history on those books, and downstream effects vary by task. The second
provider misses its combined effect-size criterion; the 305M probe covers
early training. Broader symmetric rate searches and larger models would test
transfer beyond the present settings. BF16 serving and continuous batching
would extend the fixed-batch FP32 cost measurements.

\paragraph{Conclusion.}
Local history improves in-domain prediction under shared global KV. The tested
paired-input design retains this benefit, separating content retention from
source refresh. Equal-token adaptation preserves the architecture comparison
at 8K. Complete-cache and request measurements quantify the resource
trade-off, while bounded dependencies permit exact suffix construction.

\label{sec:mainend}
\clearpage
\section*{Reproducibility statement}
Appendices~\ref{app:methods}--\ref{app:resources} provide model definitions,
data processing, evaluation protocols and proofs. The companion research archive
contains original selected-rate numerical records and scripts for recomputing
results, reference models, figure sources, and portable training and scoring
entry points. Its README and validation records distinguish the tested
short-run, restart and original-checkpoint paths from full reconstruction
instructions. Weights and source text are excluded; full raw-data reconstruction
and endpoint retraining with the portable adapter have not been rerun.
Appendix~\ref{app:artifacts} describes the artifact's coverage.

\section*{AI use statement}
Codex assisted with hypotheses, experiment design, mathematical arguments,
implementation, data checks, result analysis, literature review, writing and
figures. The author is responsible for verifying the work and its disclosure.
No synthetic language-model training text was generated for these runs.

\clearpage
\label{sec:referencesstart}
\bibliography{references}
\bibliographystyle{iclr2027_conference}
\clearpage
\appendix

\let\supplementtable\table
\renewcommand{\table}[1][H]{\supplementtable[H]}
\let\supplementfigure\figure
\renewcommand{\figure}[1][H]{\supplementfigure[H]}
\section*{Appendix overview}
\noindent\hyperref[app:methods]{A. Derivations} (p.\,\pageref{app:methods})\quad
\hyperref[app:reproduction]{B. Reproduction protocols} (p.\,\pageref{app:reproduction})\quad
\hyperref[app:primary]{C. Primary evidence} (p.\,\pageref{app:primary})\\
\hyperref[app:extensions]{D. Controls and generality} (p.\,\pageref{app:extensions})\quad
\hyperref[app:resources]{E. Execution costs} (p.\,\pageref{app:resources})\quad
\hyperref[app:artifacts]{F. Artifact and seed index} (p.\,\pageref{app:artifacts})

Training replicates use study-local letters: A--H for the factorial and A/B for
two-seed studies. Appendix~\ref{app:seedindex} records their exact identities and
checkpoint reuse. Complete endpoint tables accompany the artifact.

\FloatBarrier
\section{Method and derivations}
\label{app:methods}
The dependency proof establishes which upper positions suffice for complete-cache construction. The normalization identities explain what the fusion and multiplicity interventions change.

\FloatBarrier
\subsection{Exact suffix execution and complete-cache equivalence}
\label{app:proof}
Index prompt positions from $1$ to $N$ and let $H^j_t$ be the full upper decoder's
output at layer $j$. Lower outputs $H^0_{1:N}$ and global KV are computed in full
and held fixed. For any suffix of outputs at layer $j$ of length $r$, the union
of their local input neighborhoods lies in the suffix of length
$\min(N,r+W-1)$ at layer $j-1$. Residual connections require only the same
positions, and normalization/feedforward operations are positionwise. Global
attention adds no dependence on upper hidden positions because its source is
the already fixed lower provider. Document isolation can only shrink this union.

Define $o_j=\min(N,1+(U-j)(W-1))$ and
$i_j=\min(N,1+(U-j+1)(W-1))$. Then $o_U=1$, $i_j=\min(N,o_j+W-1)$, and
$o_{j-1}=i_j$ for $j>1$. The algorithm begins with the last $i_1$ exact lower
outputs. Suppose its layer-$j$ inputs equal the full model on this suffix.
Pointwise normalization and projections produce identical query/key/value
representations at their requested positions. Each output in its last $o_j$
positions has all permitted local predecessors inside $i_j$, and it reads the
same global KV. Original absolute-within-document rotary positions and masks
therefore give the same attention result. Residual and feedforward operations
preserve equality, proving equality on $o_j$ and establishing the induction
hypothesis for layer $j+1$.

Since $i_j\geq\min(N,W)$, each layer also projects every local KV entry required
for its saved sliding cache. Lower/global state and document metadata are
unchanged. The final singleton hidden output gives identical prompt-end logits.
Thus both logits and all saved KV match. Induction over subsequent causal decode
steps then gives identical continuations under the same supplied tokens. This
argument concerns real arithmetic; changed floating-point matrix shapes can
change rounding, which is why cache and continuation checks remain necessary.

The same dependency union yields the sufficient retained lower-boundary horizon
$R_s=W+(U-1)(W-1)$ for reconstructing the last $W$ KV entries of every upper
layer. This bound need not be minimal for a specific model or document layout.
It assumes fixed complete global KV and the stated layer operations. Additional
cross-position mixers, changed attention windows, dropped global entries or
different cache semantics require a new dependency calculation.

\FloatBarrier
\subsection{Joint fusion and the multiplicity control}
\label{app:normalization}
For one head and query, write $Z_g=\sum_{i\in G_t}e^{s^g_{ti}}$ and
$Z_l=\sum_{i\in L_t(W)}e^{s^j_{ti}}$. Each branch contains the current position,
so both sums are positive for valid queries. Dividing the two numerator sums in
Equation~\ref{eq:joint} by their respective normalizers gives
$a=(Z_g A_g+Z_l A_l)/(Z_g+Z_l)$ and
$\alpha/(1-\alpha)=Z_l/Z_g$, proving Equation~\ref{eq:branchmass}.
The identity specifies the allocation rule; measuring branch mass would
characterize how the trained model uses it.

For the repeated-current control, $m_t$ identical current local entries give
$Z_l=m_t e^{s^j_{tt}}$ and $A_l=v^j_t$. Thus multiplicity shifts the local/global
log odds by $\log m_t$ without introducing historical values. Learned scores determine branch mass. The control therefore fixes
multiplicity while allowing a different trained attention distribution.

This identity defines the joint-versus-separate comparison in
Section~\ref{sec:normalizationresults}.
Separately normalized outputs combined with the same score-dependent $\alpha$
are exactly the joint operation. A fixed or learned score-independent mixing
weight changes the mechanism. With such a weight, repeating identical entries
within a separately normalized local branch cancels, so the repeated-current and current-only controls coincide under separate fusion. Furthermore, a separately normalized one-entry branch has
$A_l=v^j_t$ and zero dependence on its local key. The joint current-only control retains key dependence
through $\alpha$. The fusions assign different roles to matching parameter tensors.
The factorial measures this complete fusion intervention within this decoder.

\FloatBarrier
\section{Data and execution protocols}
\label{app:reproduction}
All studies use document-isolated prediction and fixed splits. Optimization and
numerical recipes are specified per study.

\FloatBarrier
\label{app:data}
\paragraph{Model configurations.}
Table~\ref{tab:modelshapes} specifies both decoder shapes. Each uses tied input
and output embeddings, document-isolated attention, within-document rotary
positions, RMSNorm and SwiGLU. The projected global provider follows the
lower half of the stack; the receiving upper half retains local KV.

\begin{table}[H]
\centering\small
\begin{tabular}{@{}lrr@{}}
\toprule
Configuration & Primary & Larger probe \\
\midrule
Parameters (Local history) & 126,248,448 & 304,662,528 \\
Lower / upper layers & 8 / 8 & 12 / 12 \\
Hidden width / FFN width & 768 / 2,048 & 1,024 / 2,816 \\
Query / KV heads & 12 / 4 & 16 / 4 \\
Head dimension & 64 & 64 \\
Vocabulary / context & 32,768 / 2,048 & 32,768 / 2,048 \\
Local window (Local history / Current only) & 128 / 1 & 128 / 1 \\
\bottomrule
\end{tabular}
\caption{Complete model dimensions. These shapes define the architecture;
individual study sections specify budgets, seeds and numerical recipes.
The repeated-current control uses the current-only control shapes with the multiplicity score correction.}
\label{tab:modelshapes}
\end{table}

\paragraph{Distinct development and reserved-test data.}
Both splits contribute 1,024 complete 2K evaluation windows. Each happens to
contain 1,809 documents; masking their 1,808 cross-document transitions leaves
$1,024\times2,048-1,808=2,095,344$ valid targets in either split. Equal evaluation
counts therefore do not imply equal text. The stored streams contain 2,097,567 and 2,097,841 tokens, respectively;
unused tails are omitted. File digests accompany the artifact.
A record-level recheck verifies disjoint document IDs, canonical-page hashes,
normalized-text hashes and sampled 64-word hashes between the splits and
between either split and the extended training pool. This exact-overlap check
retains the semantic near-duplicate limitations of the preparation pipeline.

\paragraph{Data preparation and numerical admission.}

The training source is FineWeb-Edu revision
\texttt{fc9850df}, specifically
\path{sample/10BT/002_00000.parquet} and
\path{sample/10BT/003_00000.parquet}. The source declaration records the
ODC-By-1.0 dataset license, exact object sizes and hashes. The tokenizer revision
is \texttt{caa1feb0}. Full revision identifiers are retained in the artifact. All models train from newly
initialized weights. Within each seed, the frozen order hash and valid-target
count must match across arms.

For splitting, the canonical page key is the lowercased URL netloc plus its path
with trailing slashes removed; scheme, query and fragment are omitted. Missing
keys fall back to a hash of normalized text. The first 16 hexadecimal digits of
the key's SHA-256, interpreted as an integer modulo 1,000, assign buckets 0--4
to development, 5--9 to test and the remainder to training. Normalization uses
Unicode NFKC, case folding and whitespace collapsing. Documents with duplicate
normalized text, duplicate canonical page keys, or any shared sampled normalized
64-word block are rejected globally in source order. Block candidates start
every 64 words; when there are more than 32, at most 32 evenly spaced candidates
are retained. Empty text and documents larger than 2 MiB of UTF-8 are rejected.
The filter removes sampled literal overlap; semantic near-duplicates can remain.

Accepted original text is tokenized without automatic special tokens, then
wrapped in BOS and EOS. A split is filled to its target in complete-document
units. Input windows start every 2,048 tokens; one extra token supplies the
last next-token target. Adjacent windows therefore do not repeat input positions
or prediction targets, although a boundary token appears once as a target and
once as the next input. Each batch masks targets whose document ID differs from
the corresponding input ID. The original training context and ordering remain
unchanged in all matched controls.

The PG-19 repository revision is
\texttt{b0e1f292}; the protocol additionally
records source object generations. Candidate books are ordered by numeric ID
before evaluation. Whole-book/prefix normalized hashes and consecutive 64-word
candidate-prefix spans are compared with normalized document hashes and up to
32 stored 64-word samples per accepted training document. All 967,499 accepted
training-pool documents participate in this screen. Book 356 matches and is
excluded without replacement. Because the training-side spans are sampled,
unsampled overlap can remain. Evaluated windows restart context and include
explicit valid-target masks. Bootstrap sampling uses seed 777 and resamples the
49 books jointly for each paired checkpoint comparison. Each selected book
contributes at most 32,769 tokens from its BOS--text--EOS sequence. Truncation
does not add an artificial EOS.

The 2.5B-token studies extend the training pool while preserving its original
prefix and both held-out splits. Extension candidates are screened against
full/prefix normalized bodies and every contiguous 64-word span of the same
49 book prefixes, using up to 32 sampled spans per training candidate. This
guard rejects three candidate training documents; the book set is unchanged.
The resulting pool contains 2,403,417 training documents and 2.700004B tokens
(2,407,035 records including development and test). The guard and pool
manifest hashes are bound in the data-preparation records. Both screens
address sampled literal overlap, not semantic decontamination.

\paragraph{New same-source confirmation text.}
The selected-rate study uses a separately reserved sample from
\path{sample/10BT/006_00000.parquet} at the same FineWeb-Edu revision.
The first 20,000 rows are candidates. Empty/oversized records, internal
duplicates and matches to existing or external ID, normalized body, canonical
page or sampled 64-word keys are excluded. The screen uses all 2,407,035
existing records. Accepted documents remain in source order until the stream
supplies at least $2048\cdot2048+1$ tokens: 3,658 documents supply 2,048
scored windows and 4,190,647 valid targets. Selection uses no model scores.
This confirmation sample, the earlier reserved test and fixed books retain
separate identities throughout the analyses.

\paragraph{Numerical execution.}
Experiments use PyTorch 2.13.0, CUDA 13.0 and H20-3e GPUs. Training and
inference paths are qualified separately against their reference computations.
Full-cache and logit checks use $10^{-4}$ absolute/relative element tolerances
and $10^{-6}$ mean NLL tolerance unless stated otherwise. CUDA Graph replay
preserves the captured FP32 operations bitwise, including independent ownership
of returned tensors. Copies needed to provide that ownership are timed.
Compiled, BF16-incremental, SDPA and rectangular-FlexAttention candidates
that exceeded their respective thresholds were excluded from the corresponding
accepted paths. The companion's \texttt{engineering-notes.txt}
retains the rejected-path diagnostics and stage accounting.

\subsection{126M training and numerical qualification}
\label{app:training126m}
The following recipe describes the 126M history and capacity comparisons;
Appendix~\ref{app:scale305m} specifies the second configuration's numerical path.
The frozen FineWeb-Edu pool \citep{finewebedu} contains 1,100,001,261 tokens.
Each run starts from random weights and processes a seed-specific permutation of
non-overlapping prediction windows without repetition. Within each seed, arms
share inputs, order and valid-target counts. Each executes 7,630 updates with
64 sequences globally: 1,000,079,360 input tokens per run. We use AdamW \citep{adamw} with
peak learning rate $3\times10^{-4}$, betas $(0.9,0.95)$, weight decay 0.1,
gradient clipping at one, 200 warmup updates and cosine decay to 10\% of the peak.
Two H20 GPUs per model use data parallelism, eight sequences per microbatch and
four accumulation steps per rank. Loss is normalized by the global count of valid
within-document targets. The tokenizer is Mistral-7B-v0.3's release
\citep{mistraltokenizer}; no pretrained model weights are used.

The 126M runs use qualified BF16 FlexAttention and compiled cross-entropy. Numerical
checks compare explicit-reference outputs and gradients at initial and updated
states, including global valid-token weighting against a serial reference.
Fresh-process restoration compares 256 continuous updates with 128 updates
followed by restoration to update 256, covering model, optimizer, data cursor
and per-rank random states.

Development evaluation uses 1,809 documents and 2,095,344 valid targets. FP32
token cross-entropies are accumulated in float64 with per-document sums. Final
checkpoints are independently reloaded and evaluated. These historical endpoints
use development and fixed external text; Appendix~\ref{app:holdout} specifies
the separate reserved-test evaluation of the factorial and budget studies.

\subsection{Prefill numerical and graph qualification}
\label{app:prefillqualification}
Independent small-model CPU checks cover 27 cases, including document boundaries,
cache continuation and executed shapes. On both fixed history-model checkpoints, full
cache and logit checks use FP32 element tolerances $10^{-4}$ (absolute and
relative) and first-32-prediction mean NLL difference at most $10^{-6}$.
Four development and four external-book windows at lengths 512, 1,536 and 1,792
meet these tolerances. Maximum mean NLL discrepancies are
$2.12\times10^{-7}$ and $3.19\times10^{-7}$ by seed. These tolerances specify the floating-point realization of the
exact-arithmetic proposition.

To control CPU launch overhead, we also apply CUDA Graph acceleration to
\emph{all six} routes, with separate qualification of changed inputs, masks,
continuation and returned-cache ownership. Graph replay preserves its underlying
FP32 arithmetic bitwise and does not overwrite results returned for an earlier
request. Each condition has
three warmups and 15 timing repetitions in rotated order, yielding 1,080 raw
eager/graph observations across two seeds and three lengths.

These timings omit model loading and graph creation. Exact-route graph setup
takes 0.099/0.097 seconds at length 1,792, amortizing after approximately 75/69
calls at the observed saving. Its recorded live allocation increment is
65.95 MiB per graph; allocator increments depend on creation order and co-resident
graphs. The allocation measurements include three co-resident models and six graph
pools. Explicit attention forms masked score entries; an optimized sparse
full-prefix kernel could reduce the measured advantage.

\subsection{Shared zero-shot evaluation protocol}
\label{app:taskprotocol}
All task studies use PIQA validation (1,838 examples), HellaSwag validation
(10,042) and ARC-Easy test (2,376). PIQA reports accuracy; HellaSwag and ARC-Easy
report character-normalized accuracy. Each study identifies its checkpoints.

\paragraph{Task and scoring definitions.}
We use the task templates, preprocessing and metric definitions from
\citet{lmeval}, version 0.4.13, with file hashes retained in the manifest.
Dataset revisions are \texttt{142f6d7} for \texttt{baber/piqa},
\texttt{218ec52} for \texttt{Rowan/hellaswag}, and \texttt{210d026} for
\texttt{allenai/ai2\_arc}; the manifest stores their full revisions.
The original training tokenizer is unchanged. Each prompt receives a BOS
and no appended EOS. Candidate likelihood sums next-token log probabilities
for the continuation only. Normalized accuracy divides by the original
choice's character count, following the pinned task definition. Ties select
the first choice in its original order. All candidate sequences fit the
2,048-position input budget; none required truncation. Public benchmark
overlap with the web training corpus has not been audited, so these scores
are not decontaminated generalization estimates.

\paragraph{Execution qualification.}
Candidate continuations are packed into document-isolated 2K rows with reset
rotary positions and prompt-masked scores. FP32 projections with trained
TF32x3 attention pass the $10^{-3}$ token-log-probability tolerance; the initial
BF16 candidate did not. Packing and batch-size equivalence are checked before each campaign; the
original budget-model discrepancies were at most $2.07\times10^{-5}$. Metric construction matches the pinned harness.

\paragraph{Conditional uncertainty.}
Task contrasts use 5,000 paired-example bootstrap resamples (RNG seed 916),
with seed differences averaged within each example. Descriptive 95\% intervals
are unadjusted for multiplicity and condition on the trained checkpoints;
resampling examples does not increase the training-seed sample size.
Task tables place the mean and interval in one column, in percentage points,
and retain the two training-seed differences as A/B. These conditional
intervals describe example variation, whereas the primary history study
uses training-seed intervals. The artifact retains the original numerical records.

\subsection{Shared quality--cost measurement protocol}
\label{app:costprotocol}
The architecture and history-addition campaigns use H20, FP32 IEEE scoring,
CUDA-graph prefill and eager continuation. Development NLL pools all 2,095,344
valid targets. Earlier grids cross batches 1/4/8 with prompt lengths
512/1,024/1,792 and 128 extra teacher-forced forwards; prefill already produces
the first prediction. The selected-rate study uses the six cells in Appendix~\ref{app:finallr}.
Short-output extensions use prefixes of the same
continuation. Sampling, transport, loading and graph creation are outside
request timing. Complete cache counts all lower/global/local KV and document
IDs; weights, graph buffers and workspaces are reported separately.
Training seeds are the independent model replications; timing repeats describe
execution variation. Each campaign states its checkpoints, order, repetitions
and aggregation below.

\subsection{Execution cost accounting}
Scientific token budgets exclude qualification updates. Compute matching counts
training matrix products; wall time is reported separately. GPU-hour records
distinguish in-program timers from supervisor-observed allocations and include
restoration or interrupted work where specified. Study-specific costs appear
with their results; the companion retains the earlier stage inventory.
Missing whole-job timers prevent a complete-project energy or utilization estimate.

\FloatBarrier
\section{Primary quality evidence}
\label{app:primary}
The factorial isolates history and fusion; reserved-test scoring checks new
in-domain documents; longer budgets test the allocation of training computation.

\FloatBarrier
\subsection{Eight-seed history-by-fusion study}
\label{app:normresults}

The four cells cross joint versus learned separate fusion with a current-token
or 128-position local branch. We define seed-level NLL contrasts
\begin{equation}
 D_J=J_{128}-J_1,\qquad D_S=S_{128}-S_1,\qquad I=D_J-D_S.
 \label{eq:factorial}
\end{equation}
Here $J$ and $S$ denote joint and separate fusion, the subscript gives the
local window length, and each cell symbol denotes final-checkpoint NLL. Negative $D_J$ or $D_S$ favors
history; negative $I$ means a larger history benefit with joint fusion.
The four primary comparisons are $D_J$ and $I$ on development text and fixed books; the reserved-test comparisons form a separate prespecified family.
We report prespecified 98.75\% paired-seed Student-$t$ intervals, corresponding
to a Bonferroni family of four under approximately Gaussian independent seed
contrasts. Other contrasts use descriptive 95\% intervals. Each complete training seed supplies one independent observation; document
and window losses are aggregated within that seed.

\begin{figure}[!htbp]
\centering
\includegraphics[width=\linewidth]{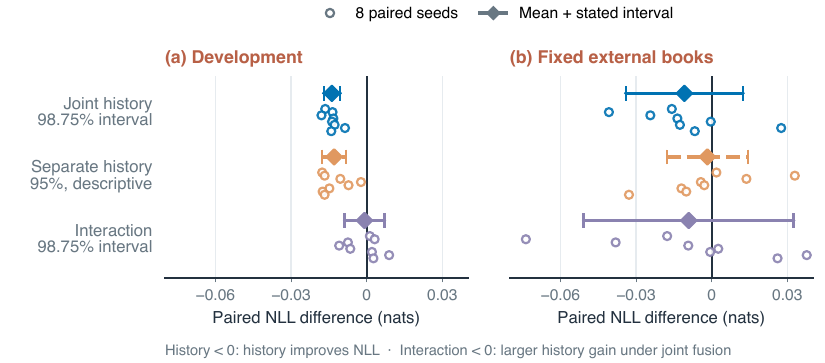}
\caption{\textbf{Local-history gains and fusion interaction are distinct.}
Thirty-two independent endpoints use 126M parameters, 1B tokens and 2K context.
Open circles show all eight paired seed effects; diamonds and bars show means
and model-based intervals at the labeled levels. Both panels share the same
scale and zero reference. Blue and orange rows show the history model minus the current-only control under joint
and separate fusion, respectively; the purple row shows the difference
between these two history effects (Equation~\ref{eq:factorial}). Negative
history values favor the history model; negative interaction means a larger history gain
under joint fusion.}
\label{fig:normalization}
\end{figure}

The four-cell design, eight seeds and final endpoints were fixed before
training. Every cell completed 7,630 updates from random initialization,
independently of the historical models and without validation-based selection. Joint
cells have 126,248,448 parameters; separate cells add 96 learned head weights.

\begin{table}[!htbp]
\centering
\begin{tabular}{lrrll}
\toprule
Study & Parameters & Seeds & Warmup & Attention path \\
\midrule
Earlier controls & 126M & A/B & 200 steps & BF16 \\
Second configuration & 305M & A/B & 200 steps & FP32 / TF32x3 \\
History $\times$ fusion & 126M & 8 new & 128 steps & FP32 / TF32x3 \\
\bottomrule
\end{tabular}
\caption{Recipe map. All rows use 1B tokens, 2K context and global batch 64;
the second row also changes depth and width. Historical discovery seed is discovery
evidence. Each inference uses matched cells within its own row.}
\label{tab:recipes}
\end{table}

The factorial uses fused AdamW \citep{adamw} with learning rate $3\times10^{-4}$,
$(\beta_1,\beta_2)=(0.9,0.95)$, weight decay 0.1 and gradient-norm clipping at
1.0. After the 128-step warmup, cosine decay reaches 0.1 of peak learning rate
at update 7,630. The learned mixing logits start at zero and use the same
parameter group. Projections/FFNs are BF16; attention uses FP32 inputs with
TF32x3 products and a compiled FP32 loss. Single-rank execution uses microbatch
32 with accumulation two. Qualified two-rank continuation uses microbatch 32
per rank with accumulation one, preserving global batch and sample order.

Fresh-process restoration is qualified against uninterrupted training, and
accepted rank changes preserve global batch and data order. Final audits
reconcile checkpoint identities, steps and targets. The recorded allocation,
including qualification and recovery, is \NormGPUHours{} GPU hours.

\begin{table}[!htbp]
\centering
\begin{tabular}{@{}lrr@{}}
\toprule
Contrast & Development & Fixed books \\
\midrule
\multicolumn{3}{@{}l}{\textit{Primary comparisons: 98.75\% seed intervals}} \\
Joint history ($D_J$) & $-13.8\;[-17.0, -10.6]$ & $-10.9\;[-34.1, +12.4]$ \\
Fusion interaction ($I$) & $-0.9\;[-8.8, +7.0]$ & $-9.1\;[-50.8, +32.5]$ \\
\addlinespace
\multicolumn{3}{@{}l}{\textit{Descriptive comparisons: 95\% seed intervals}} \\
Separate history ($D_S$) & $-12.9\;[-17.7, -8.2]$ & $-1.7\;[-17.8, +14.4]$ \\
Joint $-$ separate, history & $-18.1\;[-21.3, -15.0]$ & $-35.2\;[-50.2, -20.3]$ \\
Joint $-$ separate, current & $-17.2\;[-22.2, -12.3]$ & $-26.1\;[-51.6, -0.6]$ \\
\bottomrule
\end{tabular}
\caption{Paired-seed mean\;[interval], in $10^{-3}$ nats ($n=8$).
Negative values favor the first-named intervention; a negative interaction
means a larger history gain with joint fusion. Grouping separates the
prespecified comparisons from descriptive contrasts.}
\label{tab:normcontrasts}
\end{table}

The four primary marginal levels were prespecified. Bonferroni family
coverage uses the stated paired-seed distributional assumptions at eight seeds.
BCa intervals from 99,999 paired-seed resamples provide prespecified sensitivity
diagnostics; the primary decisions use the Student-t intervals.
All primary t/BCa pairs agree on whether zero is included. Removing any
single seed preserves a negative joint-history mean, while interaction means
can change sign
(the complete diagnostics accompany the artifact). The fixed 0.005-nat threshold
guided research continuation, not application utility or statistical significance;
intervals quantify uncertainty about the effect.

\begin{table}[!htbp]
\centering
\begin{tabular}{rlrrrr}
\toprule
Seed & Corpus & $J_1$ & $J_{128}$ & $S_1$ & $S_{128}$ \\
\midrule
A & Dev. & 3.0474 & 3.0309 & 3.0711 & 3.0534 \\
A & Books & 3.7778 & 3.7620 & 3.8137 & 3.8155 \\
\addlinespace[2pt]
B & Dev. & 3.0471 & 3.0335 & 3.0628 & 3.0461 \\
B & Books & 3.7772 & 3.7364 & 3.7430 & 3.7760 \\
\addlinespace[2pt]
C & Dev. & 3.0438 & 3.0259 & 3.0576 & 3.0471 \\
C & Books & 3.7755 & 3.7512 & 3.7946 & 3.8083 \\
\addlinespace[2pt]
D & Dev. & 3.0463 & 3.0331 & 3.0550 & 3.0527 \\
D & Books & 3.7562 & 3.7425 & 3.7805 & 3.7762 \\
\addlinespace[2pt]
E & Dev. & 3.0427 & 3.0290 & 3.0573 & 3.0501 \\
E & Books & 3.7493 & 3.7489 & 3.7905 & 3.7875 \\
\addlinespace[2pt]
F & Dev. & 3.0408 & 3.0281 & 3.0600 & 3.0451 \\
F & Books & 3.7621 & 3.7494 & 3.7808 & 3.7688 \\
\addlinespace[2pt]
G & Dev. & 3.0405 & 3.0319 & 3.0676 & 3.0500 \\
G & Books & 3.7304 & 3.7580 & 3.8062 & 3.7961 \\
\addlinespace[2pt]
H & Dev. & 3.0462 & 3.0321 & 3.0617 & 3.0450 \\
H & Books & 3.7574 & 3.7507 & 3.7852 & 3.7524 \\
\bottomrule
\end{tabular}

\caption{All 32 independently trained factorial endpoints, each evaluated on
both fixed corpora. NLLs are rounded for display; the artifact retains all published digits. No seed
is dropped or replaced; the external reversal remains in the joint-history mean.}
\label{tab:normendpoints}
\end{table}

\FloatBarrier
\subsection{Frozen reserved-test evaluation}
\label{app:holdout}
The reserved-test protocol fixes all 32 factorial and eight budget endpoints,
comparisons, aggregation rules and interval levels before scoring. Final-step
checkpoints are used throughout. The split uses the disjointness and duplicate
rules in Appendix~\ref{app:data}; it tests new in-domain documents.

All models score the first 1,024 complete 2K windows in stored order:
\HoldoutTargets{} targets across \HoldoutDocuments{} documents. Microbatch-32
evaluation first reproduces stored development NLL to $10^{-7}$ and per-document
sums to $10^{-4}$, then scores test once. An independent aggregation verifies
counts, losses and contrasts.

The factorial's two primary test contrasts are $D_J$ and $I$, with 97.5\%
paired-seed Student-$t$ intervals (Bonferroni family of two, approximately
Gaussian independent seed contrasts). Other test contrasts use descriptive
95\% intervals. This family is distinct from the development/book family;
training seeds remain the independent units.

\begin{table}[ht]
\centering
\begin{tabular}{@{}lrr@{}}
\toprule
Contrast & Mean\;[interval] & Negative seeds \\
\midrule
\multicolumn{3}{@{}l}{\textit{Primary comparisons: 97.5\% seed intervals}} \\
Joint history ($D_J$) & $-14.2\;[-17.3, -11.1]$ & 8/8 \\
Fusion interaction ($I$) & $-1.2\;[-7.8, +5.3]$ & 3/8 \\
\addlinespace
\multicolumn{3}{@{}l}{\textit{Descriptive comparisons: 95\% seed intervals}} \\
Separate history ($D_S$) & $-12.9\;[-17.9, -7.9]$ & 8/8 \\
Joint $-$ separate, history & $-18.6\;[-21.7, -15.5]$ & 8/8 \\
Joint $-$ separate, current & $-17.4\;[-22.4, -12.3]$ & 8/8 \\
\bottomrule
\end{tabular}
\caption{Reserved-test paired-seed contrasts, in $10^{-3}$ nats ($n=8$).
Negative-seed counts preserve the direction of each independent training pair.}
\label{tab:testcontrasts}
\end{table}

Budget-test results retain both paired values and their descriptive mean
(Table~\ref{tab:budgetcontrasts}), with no training-seed population interval.
Test scores do not select models or additional seeds.

\FloatBarrier
\subsection{Longer budgets and matched training computation}
\label{app:budgetprotocol}
\begin{figure}[!ht]
\centering
\includegraphics[width=\linewidth]{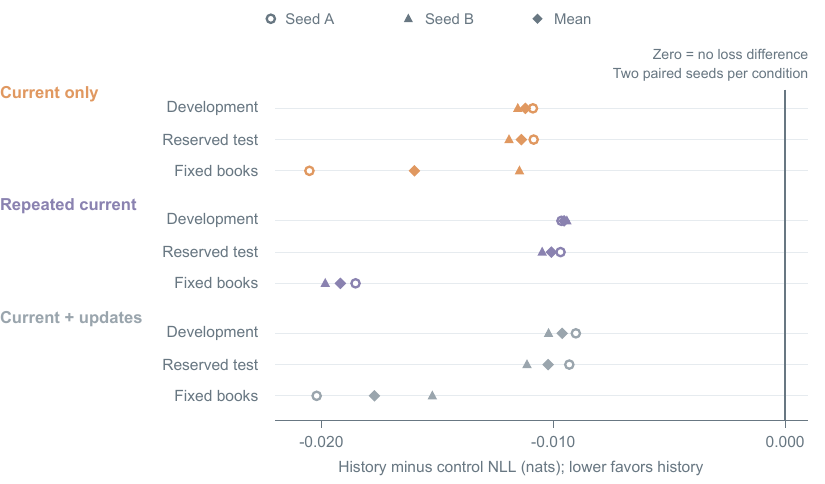}
\caption{\textbf{History retains its value under longer and matched training budgets.}
At 126M and 2K context, circles/triangles show paired seeds A/B; diamonds show means. Negative NLL
differences favor the history model. The current-only control and the repeated-current control share its 2.5B-token budget; the compute-matched current-only control receives
matched matrix-product FLOPs. Points show seed variation, not confidence intervals.}
\label{fig:budget}
\end{figure}
The common-budget study fixes the history model, the current-only control and the repeated-current control at 126M parameters and
2K context, using paired seeds A/B. Each model trains
from random initialization for 19,074 updates, or 2,500,067,328 input tokens,
at global batch 64. The training pool extends the earlier pool while
preserving its prefix and the development and test splits. All arms use a
peak learning rate of 0.0003, a 128-update warmup and a cosine schedule spanning
their full budget.

For each seed, a fresh compute-matched current-only control uses 19,267 updates
(2,525,364,224 input tokens) and a cosine schedule spanning its own budget.
Its counted training matrix-product FLOPs exceed the history model's by less than 0.001\%.
The count includes forward and backward matrix products under the actual
document-isolated attention windows; one multiply-add counts as two FLOPs.
Softmax, normalization, rotary transforms, elementwise operations, optimizer
updates, kernel padding and recomputation are outside this convention.
This is a training-computation comparison; it does not match elapsed time
or inference latency.

Final checkpoints are independently reloaded to verify scores and valid-target
counts. Qualified continuation preserves global batch, data order and the
optimization schedule; qualification updates are excluded from token budgets.

\begin{table}[ht]
\centering\small
\caption{NLL differences in $10^{-3}$ nats. Local history minus each comparator. Seeds A/B follow the common-budget pairing. Negative $\Delta$ favors history. Percentages use the geometric-mean PPL ratio, not the mean of two percentages.}
\label{tab:budgetcontrasts}
\begin{tabular}{llrrr}
\toprule
Condition & Comparator & Seed A $\Delta$ & Seed B $\Delta$ & PPL decrease \\
\midrule
Development & Current only & -10.9 & -11.5 & 1.11\% \\
 & Repeated current & -9.6 & -9.4 & 0.95\% \\
 & Current + updates & -9.0 & -10.2 & 0.96\% \\
\addlinespace[2pt]
\addlinespace[3pt]
Reserved test & Current only & -10.8 & -11.9 & 1.13\% \\
 & Repeated current & -9.7 & -10.5 & 1.00\% \\
 & Current + updates & -9.3 & -11.1 & 1.02\% \\
\addlinespace[2pt]
\addlinespace[3pt]
Fixed books & Current only & -20.5 & -11.5 & 1.59\% \\
 & Repeated current & -18.5 & -19.8 & 1.90\% \\
 & Current + updates & -20.2 & -15.2 & 1.76\% \\
\addlinespace[2pt]
\bottomrule
\end{tabular}
\end{table}

NLL is aggregated by valid target count within each checkpoint and condition.
Within-seed differences are then averaged equally across the two seeds;
exponentiating this mean gives the geometric-mean perplexity ratio. Both
paired differences are reported. These results characterize the fixed budget
and recipe, while the eight-seed factorial supplies the broader seed comparison.

\FloatBarrier
\section{Controls and generality checks}
\label{app:extensions}
Capacity, entry count, configuration, provider and downstream comparisons each
retain their own recipe, seeds and inferential scope.

\FloatBarrier
\subsection{Equal learning-rate search and independent confirmation}
\label{app:finallr}
This study separates recipe selection from architecture confirmation. Local
history, GQA2 and \CLABaseline{} receive the same learning-rate grid
$\{1.5,3,6\}\times10^{-4}$ on selection seed (common A). Final development NLL selects
each rate, with ties assigned to the lower rate. Three compatible center-rate
endpoints are reused; six new runs complete the grid. All models select
$6\times10^{-4}$ (Table~\ref{tab:finallr}). The frozen rule stops at this
grid, including when its best observed point is on the boundary.

\begin{table}[H]
\centering\small
\caption{Equal three-point learning-rate search on selection seed (common A). Entries are final development NLL. The center-rate endpoints are reused; all three selected rates lie at the upper boundary.}
\label{tab:finallr}
\begin{tabular}{@{}lrrr@{}}
\toprule
Architecture & $1.5\times10^{-4}$ & $3\times10^{-4}$ & $6\times10^{-4}$ \\
\midrule
Local history & 2.9420 & 2.8257 & 2.7754 \\
GQA2 & 2.9662 & 2.8506 & 2.7942 \\
GQA4--CLA2 & 2.9688 & 2.8455 & 2.7858 \\
\bottomrule
\end{tabular}
\end{table}

\paragraph{Confirmation and data separation.}
Six fresh models use confirmation seeds A/B, paired across architectures.
Architecture, packed order, global batch 64 and matrix-product budget matching
follow the earlier comparison. Local history takes 19,074 updates (2.5001B input
tokens); the alternatives take 19,269 matched-compute updates. All configuration
choices are frozen before confirmation. Final weights are independently reloaded
and all six checkpoint hashes bound before scoring reserved confirmation text.
This text has 2,048 windows and 4,190,647 valid targets from FineWeb-Edu;
it was not used for learning-rate selection and is distinct from the earlier
reserved test. The fixed 49-book condition has 1,475,654 valid targets.
It is unchanged from earlier studies and is not a fresh external test.

\begin{table}[H]
\centering\small
\caption{All selected-rate confirmation endpoints. A/B are the confirmation seeds; lower NLL is better. Confirmation text is a reserved sample from the training source, separate from the earlier reserved test.}
\label{tab:finalquality}
\begin{tabular}{@{}llrrr@{}}
\toprule
Seed & Architecture & Development & Confirmation & Fixed books \\
\midrule
A & Local history & 2.7748 & 2.8743 & 3.3946 \\
A & GQA2 & 2.7880 & 2.8893 & 3.4764 \\
A & GQA4--CLA2 & 2.7882 & 2.8881 & 3.4460 \\
B & Local history & 2.7805 & 2.8803 & 3.4203 \\
B & GQA2 & 2.7932 & 2.8941 & 3.4829 \\
B & GQA4--CLA2 & 2.7897 & 2.8895 & 3.4500 \\
\bottomrule
\end{tabular}
\end{table}

\paragraph{Likelihood and interpretation.}
Reported reductions are $100[1-\exp(\overline{\Delta\mathrm{NLL}})]$, averaging
paired differences over the two training seeds. Against GQA2 and \CLABaseline{},
confirmation reductions are \FinalConfirmGQA\% and \FinalConfirmCLA\%; fixed-book
reductions are \FinalBooksGQA\% and \FinalBooksCLA\%. Each individual pair favors
history in both conditions. These are new-seed comparisons of selected
configurations, not training-population confidence intervals. The common-recipe
and selected-rate studies differ in both learning rate and confirmation seeds;
their difference does not isolate a causal effect of tuning. The eight-seed
history-versus-current study answers a separate, within-backbone question.

\paragraph{Fixed downstream tasks.}
The six checkpoints use the existing PIQA, HellaSwag and ARC-Easy templates,
FP32 scoring and metrics, with no task-based model selection. Packed and batched
scoring equivalence is qualified before scoring. Benchmark overlap with the
training corpus has not been audited. Both seeds favor the alternatives on
PIQA and history on HellaSwag; ARC-Easy reverses direction between seeds.
Table~\ref{tab:finaltaskintervals} reports both seed differences and their intervals;
the artifact retains all absolute accuracies.
All six conditional intervals include zero, leaving a general task-accuracy
advantage unestablished.

\begin{table}[H]
\centering
\begin{tabular}{@{}llrrr@{}}
\toprule
Task & Comparator & A & B & Mean\;[95\% CI] \\
\midrule
PIQA & GQA2 & $-1.09$ & $-0.22$ & $-0.65\;[-1.69, +0.38]$ \\
 & GQA4--CLA2 & $-0.71$ & $-0.05$ & $-0.38\;[-1.44, +0.68]$ \\
\addlinespace[3pt]
HellaSwag & GQA2 & $+0.25$ & $+0.27$ & $+0.26\;[-0.13, +0.66]$ \\
 & GQA4--CLA2 & $+0.30$ & $+0.33$ & $+0.31\;[-0.09, +0.71]$ \\
\addlinespace[3pt]
ARC-Easy & GQA2 & $+0.72$ & $-0.42$ & $+0.15\;[-0.93, +1.22]$ \\
 & GQA4--CLA2 & $+1.26$ & $-0.42$ & $+0.42\;[-0.57, +1.41]$ \\
\bottomrule
\end{tabular}
\caption{Selected-rate task differences, local history minus comparator
(percentage points). The protocol and metrics are in Appendix~\ref{app:taskprotocol}.
A/B are training-seed differences; intervals condition on the fixed checkpoints.}
\label{tab:finaltaskintervals}
\end{table}

\paragraph{Costs of the same selected checkpoints.}
Two fixed shapes $(B,N)=(1,1792),(4,1024)$ are crossed with $D=1,8,128$
extra teacher-forced decode steps. All retain complete caches. FP32 IEEE
development scoring supplies Figure~\ref{fig:gqatradeoff}'s quality coordinate;
Table~\ref{tab:finalquality} instead uses the qualified training evaluation path.
The small precision differences are kept separate. Each checkpoint passes
91 numerical checks covering full/exact cache construction, 128-step continuation,
packed documents and state restoration. Three timing blocks rotate model order
on the same physical GPU within each seed. Four retained repeats per route,
cell and block give 864 timing records. Per-block medians are combined through
geometric mean paired ratios. Blocks and repeats do not increase the seed count.
Captured prefill and eager decode measure a fixed-batch reference workload.

\begin{table}[H]
\centering\small
\caption{All six prespecified request cells. Ratios are local history divided by the comparator, geometrically averaged over three rotated timing blocks and two seeds. Values below one favor local history. Cache is complete post-prefill storage, including document IDs.}
\label{tab:finalcost}
\begin{tabular}{@{}llrrr@{}}
\toprule
$B/N/D$ & Comparator & Prefill & Request & Cache \\
\midrule
1/1792/1 & GQA2 & 0.775 & 0.857 & 1.196 \\
1/1792/1 & GQA4--CLA2 & 0.835 & 0.939 & 1.196 \\
1/1792/8 & GQA2 & 0.776 & 1.033 & 1.196 \\
1/1792/8 & GQA4--CLA2 & 0.837 & 1.176 & 1.196 \\
1/1792/128 & GQA2 & 0.777 & 1.137 & 1.196 \\
1/1792/128 & GQA4--CLA2 & 0.836 & 1.330 & 1.196 \\
4/1024/1 & GQA2 & 0.925 & 0.960 & 1.250 \\
4/1024/1 & GQA4--CLA2 & 0.992 & 1.041 & 1.250 \\
4/1024/8 & GQA2 & 0.926 & 1.058 & 1.250 \\
4/1024/8 & GQA4--CLA2 & 0.992 & 1.187 & 1.250 \\
4/1024/128 & GQA2 & 0.924 & 1.148 & 1.250 \\
4/1024/128 & GQA4--CLA2 & 0.990 & 1.325 & 1.250 \\
\bottomrule
\end{tabular}
\end{table}

\begin{figure}[H]
\centering
\includegraphics[width=\linewidth]{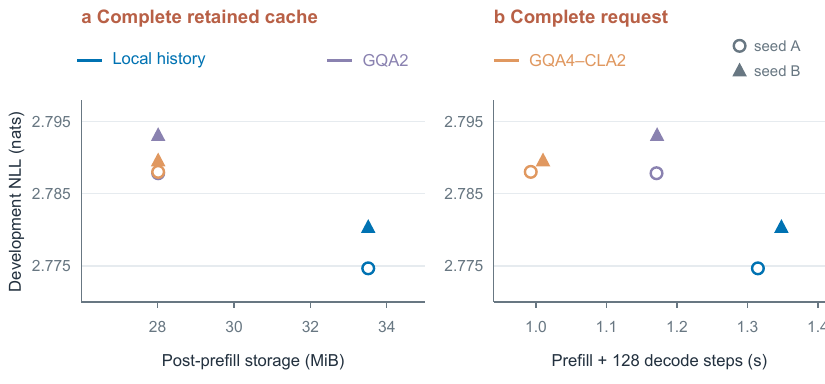}
\caption{\textbf{Absolute quality--resource operating points for long requests.}
The six selected-rate 126M-scale checkpoints supply FP32 development NLL, complete
post-prefill cache and request time; lower is better on both axes. Circles
and triangles identify the two confirmation seeds. Batch one, a 1,792-token
prompt and 128 extra teacher-forced decode steps use graph prefill and eager
decode. Times aggregate three rotated blocks. Figure~\ref{fig:gqatradeoff}
compares request ratios across all six workloads.}
\label{fig:requestduration}
\end{figure}

At batch one, the 5.5 MiB cache premium comprises a 3.5 MiB full-length bank
and 2 MiB of local entries. Relative to \CLABaseline{}, prefill is about
\FinalPrefillSaving\% shorter, but the 128-step request is \FinalRequestPremium\%
longer. The one-step request favors history at batch one; the eight-step request
favors \CLABaseline{}, which also wins at batch four with one step.
The measured trade-off depends on output length and batch size, rather than
establishing a general request-speed advantage. Earlier 15-cell measurements
in Appendix~\ref{app:clacommon} use their own original checkpoints.

\FloatBarrier
\subsection{Equal-token adaptation to 8K context}
\label{app:contextadaptation}
This study asks whether the whole-design quality ordering persists after both
models adapt to a longer context. It continues the selected-rate local-history
and \CLABaseline{} checkpoints from Appendix~\ref{app:finallr}, using the same
two confirmation seeds A/B. The comparison therefore extends
these models in context length, rather than adding independent training seeds.

\paragraph{Common adaptation recipe.}
Both models use Position Interpolation \citep{positioninterpolation}: rotary
positions are divided by four, while causal visibility and the 128-token local
window retain their token units. Each model takes 1,024 updates of 16 sequences
of length 8,192, totaling 134,217,728 tokens. AdamW starts from fresh optimizer
state with $(\beta_1,\beta_2)=(0.9,0.95)$, weight decay 0.1 and gradient clipping
at 1. The shared peak learning rate is $10^{-4}$, with 64 warmup steps and
cosine decay to 10\% of the peak. This fixed recipe matches adaptation tokens;
it does not match adaptation FLOPs or perform a new rate search. Final-step
checkpoints are used throughout.

The original frozen training pool supplies 36,765 nonoverlapping 8K prediction
windows from 25,078 long documents. Each adaptation uses 16,384 of these windows
without repetition, with identical ordering within a seed pair. These are
existing training documents, not newly collected text. No development or book
text enters adaptation. Training uses FP32 projections, FP32 attention with
TF32x3 products, fused loss and AdamW, and a jointly qualified tiled backward
kernel. The BF16 projection candidate failed the prespecified output tolerance
at 8K and was excluded before formal training; the FP32 path passed the same
output, loss and all-parameter gradient criteria. All four configurations also
passed bitwise continuous-16 versus fresh-process-8+8 update comparisons.

\paragraph{Matched prediction targets.}
The long same-source condition contains all 30 eligible 8K development windows
from 19 documents. The book condition uses the original 49 fixed PG-19 books.
For a length-$L$ forward, inputs are positions $8192-L,\ldots,8191$ of an
8,193-token sequence; all lengths score targets $7169,\ldots,8192$.
Thus 2K, 4K and 8K compare the same 1,024 predictions with different available
prefixes: 30,720 targets on long development text and 50,176 on books per
checkpoint. Both conditions are reused evaluation text; this is an exploratory
follow-up, not a new held-out confirmation. The original 2K regression test
uses all 1,024 development windows and 2,095,344 within-document targets.
Scoring uses FP32 projections and IEEE dense reference attention.

Figure~\ref{fig:contextadaptation} in the main text plots the paired architecture
advantage at each length. For seed $s$, let $\Delta_s=\ell_{\mathrm{history},s}
-\ell_{\mathrm{CLA},s}$. Each seed point is $100(1-\exp\Delta_s)$; the blue
summary is $100[1-\exp(\operatorname{mean}_s\Delta_s)]$, rather than the
arithmetic mean of seed percentages. The shared zero-based vertical scale and
logarithmic context axis show the measured lengths without extrapolation.
Individual seed lines describe checkpoint variation, not confidence bounds.

\begin{table}[H]
\centering\small
\caption{\textbf{Likelihood after equal-token 8K adaptation.} Each row scores the same final 1,024 targets per sequence; 30,720 targets for long development text and 50,176 for books. Lower NLL is better. $\Delta$PPL is local history relative to GQA4--CLA2; negative values favor history. A/B are the confirmation seeds.}
\label{tab:contextquality}
\begin{tabular}{@{}llrrrr@{}}
\toprule
Text & Context / seed & Local NLL & CLA NLL & $\Delta$NLL & $\Delta$PPL (\%) \\
\midrule
Long dev. & 2K / A & 2.7399 & 2.7589 & -0.01902 & -1.88 \\
Long dev. & 2K / B & 2.7505 & 2.7550 & -0.00451 & -0.45 \\
Long dev. & 4K / A & 2.7134 & 2.7308 & -0.01743 & -1.73 \\
Long dev. & 4K / B & 2.7239 & 2.7285 & -0.00463 & -0.46 \\
Long dev. & 8K / A & 2.6964 & 2.7139 & -0.01749 & -1.73 \\
Long dev. & 8K / B & 2.7073 & 2.7132 & -0.00592 & -0.59 \\
Fixed books & 2K / A & 3.3439 & 3.3740 & -0.03013 & -2.97 \\
Fixed books & 2K / B & 3.3840 & 3.3959 & -0.01194 & -1.19 \\
Fixed books & 4K / A & 3.3117 & 3.3447 & -0.03296 & -3.24 \\
Fixed books & 4K / B & 3.3571 & 3.3700 & -0.01282 & -1.27 \\
Fixed books & 8K / A & 3.2973 & 3.3402 & -0.04297 & -4.21 \\
Fixed books & 8K / B & 3.3476 & 3.3637 & -0.01601 & -1.59 \\
\bottomrule
\end{tabular}
\end{table}

At 8K, the geometric-mean perplexity reductions relative to \CLABaseline{}
are \ContextLongGain\% on long development text and \ContextBookGain\% on books;
both seeds favor local history. Increasing available context from 2K to 8K
lowers the adapted history model's book perplexity by 4.06\%, versus 3.25\%
for CLA. On long development text, the corresponding changes are nearly the
same, 4.24\% and 4.25\%. The retained ordering and the gain from extra context
are distinct findings; these observations do not establish a uniformly larger
context benefit for one architecture.

\begin{table}[H]
\centering\small
\caption{\textbf{The 2K cost of the shared adaptation recipe.} All columns score the same 2,095,344 valid development targets. Native uses the original positions and parent weights; PI before and after use positions divided by four. The last column compares adapted PI with native, without changing architecture.}
\label{tab:contextshort}
\begin{tabular}{@{}llrrrr@{}}
\toprule
Model & Seed & Native NLL & PI before & PI after & $\Delta$PPL (\%) \\
\midrule
Local history & A & 2.7747 & 4.3746 & 2.8021 & +2.78 \\
Local history & B & 2.7804 & 4.4274 & 2.8075 & +2.75 \\
GQA4--CLA2 & A & 2.7880 & 4.3965 & 2.8152 & +2.76 \\
GQA4--CLA2 & B & 2.7896 & 4.1783 & 2.8173 & +2.81 \\
\bottomrule
\end{tabular}
\end{table}

\paragraph{Short-context retention and interpretation.}
The common adaptation raises original 2K development perplexity by 2.76\%
for local history and 2.78\% for CLA, averaging paired NLL changes before
exponentiation (Table~\ref{tab:contextshort}). The history model remains
\ContextShortGain\% better than CLA on that condition. Thus the architecture
ordering survives adaptation, alongside a shared short-context cost.
All four endpoints are independently reloaded before scoring; every length,
seed and condition is retained. Two seeds describe these checkpoints, while
the eight-seed history-versus-current experiment supplies the separate
within-backbone inference. No 8K serving-speed claim follows from these scores.

\paragraph{Unadapted extension and interpolation-only checks.}
Before adaptation, a separate full-forward diagnostic evaluated the native
2K models on these same book targets. Mean NLL at 2K/4K/8K was
3.357/4.971/6.314 for local history and 3.394/7.494/7.741 for CLA;
both degraded when extended without adaptation. FP32 and FP64 aggregate
scores agreed within $7.3\times10^{-7}$ nats. The smaller deterioration of
one model is not evidence of a successfully adapted long-context model.
Interpolation-only parent scores are also retained in the accompanying records;
Table~\ref{tab:contextshort} shows the 2K case. This diagnostic motivated the
common adaptation study, rather than selecting a favorable length endpoint.
The earlier cached-continuation timing precheck did not meet its numerical
tolerance and supplies no accepted long-context timings.

\FloatBarrier
\subsection{Capacity and local-content controls}
\label{app:historical}
\label{sec:quality}
\label{sec:scale305m}
These controls separate local content from parameter count and entry
multiplicity. The 305M early-training probe and alternative global provider
are reported in Appendices~\ref{app:scale305m} and~\ref{app:providerextension},
including the latter's unmet combined effect-size criterion.

\paragraph{Training controls.}
The independent-KV baseline uses independent full causal KV in every layer. The global-only control uses the shared provider
without upper local KV. The history model adds each upper
layer's own 128-position local branch. The current-only control keeps all history-model parameter shapes but
restricts the local window to the current position, preserving access to
history through shared global KV. The pointwise control adds a rank-256 pointwise residual
adapter after each upper block of the global-only control,
$x\leftarrow x+W_{\rm down}\operatorname{SiLU}(W_{\rm up}\RMS_{n2}(x))$,
reusing its second normalization and adding no temporal state. The pointwise control and the history model have
exactly 126,248,448 parameters. This matches parameter count; their FLOPs and
latencies differ. Identically named parameters use seed-and-name keyed
initialization, so common tensors start identically within each seed.

These earlier 126M comparisons test capacity, local content and multiplicity.
Their two confirmation seeds are distinct from the eight-seed factorial.
\begin{table}[H]
\centering\small
\begin{tabular}{@{}lrrrrr@{}}
\toprule
& & \multicolumn{2}{c}{Development NLL} & \multicolumn{2}{c}{Book NLL} \\
\cmidrule(lr){3-4}\cmidrule(l){5-6}
Model & Params. (M) & A & B & A & B \\
\midrule
Independent KV & 125.854 & 3.0406 & 3.0464  & --- & --- \\
Global only & 123.103 & 3.0476 & 3.0503  & 3.7813 & 3.7505 \\
Global + adapters & 126.248 & 3.0419 & 3.0459  & --- & --- \\
Local history & 126.248 & 3.0286 & 3.0278  & 3.7473  & 3.7298  \\
Current only & 126.248 & 3.0396 & 3.0440  & 3.7499 & 3.7385 \\
Repeated current & 126.248 & 3.0435 & 3.0449  & 3.7764 & 3.7585 \\
\bottomrule
\end{tabular}
\caption{Earlier 126M controls at 1B tokens. A/B are paired confirmation runs;
book scores were collected for the four listed local/global controls.
The four discovery endpoints are retained in the artifact, separately from confirmation.}
\label{tab:dev}\label{tab:external}
\end{table}

\begin{figure}[!htbp]
\centering
\includegraphics[width=\linewidth]{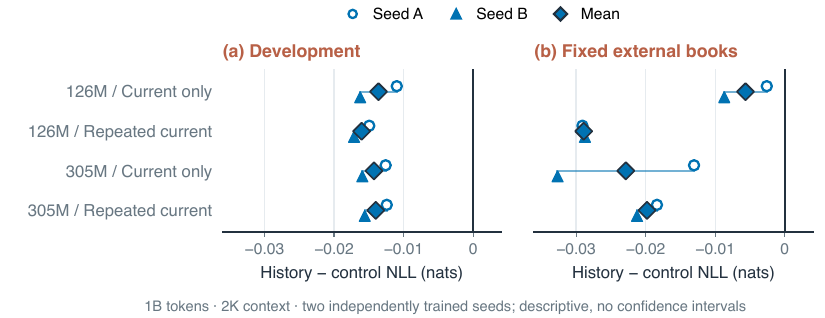}
\caption{Local-history contrasts at the fixed 1B-token endpoint. Rows label the comparator; every value is the history model minus that independently
trained control. Both seeds and their mean are shown.
Marker shapes distinguish the seeds and mean; no confidence interval is shown. Negative
values favor the history model. Both configurations use 2K context and 1B training tokens; the 305M
configuration changes depth, width and numerical path.}
\label{fig:quality}
\end{figure}

Both seeds favor history on development text (Figure~\ref{fig:quality}).
On fixed books, history improves over current-only on 34/49 and 41/49 books;
the conditional 95\% intervals are $[-6.5,+1.6]$ and $[-13.5,-3.4]$ in
$10^{-3}$ nats. The first pair remains unresolved. Against repeated-current,
both seeds improve on 48/49 books, with intervals $[-32.2,-25.8]$ and
$[-32.9,-25.3]$ in the same units. These intervals use 2,000 paired-book
resamples and condition on each checkpoint pair. The artifact retains the
original contrasts and timing records.

\FloatBarrier
\paragraph{Multiplicity validation and decision rule.}
\label{app:pending}
The repeated-current control fixes seeds A/B, the original 16-layer configuration, initialization,
data order, optimizer, global batch and 1B-token endpoint. Literal-duplicate
CPU output/gradient, causality, document-isolation and $m_t=1$ checks pass,
as do full-width BF16, compiled-loss, global-gradient and fresh-process
restoration checks. Endpoint evaluation separately measures predictive quality.

Before the runs, the historical-content continuation rule required both the history model
minus the repeated-current control development-NLL contrasts to be negative and their mean at most
$-0.005$. The current-only control comparison uses the same sign and mean-difference threshold.
The completed results meet the prespecified engineering continuation
criteria. Statistical inference is reported separately; seeds, endpoints and
thresholds retain their original definitions. Independent audits reconciled checkpoint identities,
source/config hashes, data order, valid-target counts and all 1,809 document
sums per model. Both endpoints were reloaded for the unchanged 49-book secondary
condition; these historical runs did not score the reserved test.

\FloatBarrier
\subsection{305M configuration probe}
\label{app:scale305m}
The probe uses the 24-layer configuration in Table~\ref{tab:modelshapes},
with 12 lower and 12 upper layers, hidden width 1,024 and FFN width 2,816.

\begin{table}[!htbp]
\centering
\begin{tabular}{llrrr}
\toprule
Condition & NLL contrast & A & B & Mean \\
\midrule
Development & Local history$-$Current only & $-12.6$ & $-15.9$ & $-14.3$ \\
 & Local history$-$Repeated current & $-12.4$ & $-15.6$ & $-14.0$ \\
\addlinespace[3pt]
Fixed books & Local history$-$Current only & $-13.0$ & $-32.7$ & $-22.9$ \\
 & Local history$-$Repeated current & $-18.4$ & $-21.2$ & $-19.8$ \\
\bottomrule
\end{tabular}

\caption{NLL differences in $10^{-3}$ nats. 305M paired NLL differences at 1B tokens. Both controls and all six
endpoints are reported; negative values favor local history.}
\label{tab:scale305m}
\end{table}

\begin{table}[!htbp]
\centering
\begin{tabular}{@{}llrr@{}}
\toprule
Contrast & Seed & 95\% CI & Books improved \\
\midrule
Local history$-$Current only & A & $[-17.4,-8.9]$ & 40/49 \\
Local history$-$Repeated current & A & $[-21.6,-15.5]$ & 48/49 \\
Local history$-$Current only & B & $[-36.0,-29.5]$ & 49/49 \\
Local history$-$Repeated current & B & $[-24.6,-18.0]$ & 48/49 \\
\bottomrule
\end{tabular}

\caption{NLL differences in $10^{-3}$ nats. 305M paired-book 95\% percentile intervals from 2,000 resamples of
the same 49 books. These condition on each trained checkpoint pair; they are
neither training-seed intervals nor simultaneous coverage for four contrasts.}
\end{table}

Every endpoint executes 7,630 scientific updates, or 1,000,079,360 input tokens,
on the same no-repeat training pool. Within each seed, all three arms share
the named-tensor initialization, sample permutation, valid-target counts and
schedule. The independent audit reconciles the complete segmented trajectories,
all development document sums and checkpoint identities. External evaluation
reaggregates all 4,350 model-window and 294 model-book records. These 305M
endpoints are outside the frozen 126M reserved-test evaluation. The frozen engineering rule, requiring both history minus repeated-current development
contrasts negative and their mean at most $-0.005$, passes. This engineering criterion is separate from significance testing;
all external endpoints follow the fixed evaluation plan.

The 305M recipe uses FP32/TF32x3 attention, BF16 projections and compiled loss.
Backward tiling is qualified before use; interrupted runs resume from verified
states without changing budgets. Allocation, including recovery and audits,
is \ScaleTrainHours{} GPU hours, plus \ScaleExternalHours{} for external scoring.
The artifact retains stage-level accounting and numerical checks.

\FloatBarrier
\subsection{A second global provider}
\label{app:providerextension}
The alternative provider shares post-RoPE keys from lower block eight and
values from lower block one across all upper layers, with gradients flowing
into the original projections. Removing the boundary normalization and KV
projection gives 125,854,464 parameters in each arm. Cross-provider comparisons
are descriptive; within-provider arms have equal parameter counts.

\paragraph{Fixed design and execution.}
The history model, the current-only control and the repeated-current control each train with seeds A/B, 2K contexts and global
batch 64 for 7,630 updates (1,000,079,360 input tokens). They share named-tensor
initialization and data order within each seed, using the original 1B manifest.
AdamW uses peak learning rate $3\times10^{-4}$, 200 warmup updates, cosine decay
to one tenth of peak, $(\beta_1,\beta_2)=(0.9,0.95)$, weight decay 0.1 and
clipping at 1.0. All arms use BF16 projections, FP32 attention with TF32x3
products and compiled FP32 cross-entropy. This common precision amendment
follows a rejected BF16 gradient qualification without relaxed tolerances.
The design predates the reserved-test evaluation; GPU implementation follows
it. The original final test is not reopened.

The common implementation passes output, gradient, global-batch and bitwise
restart checks; the BF16 candidate is excluded before training. Final reloads
verify development scores before book scoring. Training and verification use
35.15 GPU hours, with another 0.04 for books.

\paragraph{Results and prespecified decision.}
The history model improves on both controls on both seeds in both conditions
(Table~\ref{tab:providercontrasts}). Against the current-only control, mean development NLL changes
by $-0.009504$ nats, corresponding to 0.95\% lower geometric-mean perplexity;
against the repeated-current control, it changes by $-0.003900$ nats (0.39\%). Fixed-book reductions
are 2.28\% and 2.56\%, respectively. The preset engineering criterion requires
both development differences to be negative and their mean to be at most
$-0.005$ nats for each control. The current-only control contrast meets that criterion; the repeated-current control
contrast does not, so the combined criterion is not met. The directional
benefit extends to this provider, while the increment beyond the entry-count
control is smaller than the prescribed threshold.

\begin{table}[!htbp]
\centering
\begin{tabular}{llrrr}
\toprule
Control & Seed & Dev. & Books & Book 95\% CI \\
\midrule
Current only & A & $-13.4$ & $-10.4$ & $[-14.4,-7.1]$ \\
 & B & $-5.6$ & $-35.7$ & $[-39.8,-32.1]$ \\
\addlinespace[3pt]
Repeated current & A & $-4.8$ & $-10.5$ & $[-15.5,-6.4]$ \\
 & B & $-3.0$ & $-41.4$ & $[-46.6,-36.7]$ \\
\bottomrule
\end{tabular}

\caption{NLL differences in $10^{-3}$ nats. History minus each control with the alternate provider. Book
intervals use 2,000 paired resamples of the 49 books, RNG seed 777, retaining
token weighting. They condition on fixed checkpoints and do not estimate
training-seed uncertainty.}
\label{tab:providercontrasts}
\end{table}

All four conditional book intervals exclude zero. These two-seed results
extend the directional finding; the separate eight-seed book inference remains unresolved.

\FloatBarrier
\subsection{Sharing the source of local representations}
\label{app:localsource}
The source-by-history comparison asks whether retaining historical content is
useful only when every layer forms it from a new input. We cross per-layer and
adjacent-pair inputs with 128-position history and repeated-current entries. A
further history variant shares the projected local KV bank.

\paragraph{Input sources and projected banks.}
The eight upper layers form four adjacent pairs. The history model normally forms local KV
from each receiver's input. \emph{Shared input} instead feeds the pair's entry
representation to both receivers' separate RMSNorm and KV projections,
preserving 126,248,448 parameters. In its repeated-current control, each receiver uses this
shared input's current-token entry, with the same-document $+\log m_t$ term
from Appendix~\ref{app:historical}. The control matches entry count, not learned
attention mass or training computation. \emph{Shared local KV} also reuses the
projected bank, removing the second KV projection: 124,675,584 parameters and
about 1.104\% fewer counted training matrix products. All variants retain their
queries, residual streams, feedforward blocks and global provider. Source
sharing changes current and historical features together, including gradient
paths; sharing projected KV also changes capacity and computation.

\paragraph{Matched training and scoring.}
Seeds A/B follow the common-budget pairing. Every endpoint uses the same longer-budget
data, tokenizer, initialization convention, AdamW recipe and 2K/global-64
updates: 19,074 updates (2,500,067,328 input tokens), peak learning rate
$3\times10^{-4}$, 128 warmup steps and cosine decay to one tenth of peak.
The original source variants were fixed before their results. The two
shared-input repeated-current cells were specified prospectively after those results and
the PLT comparison; the full four-cell study was not preregistered together.
Independent reloads reproduce development scores; external scoring retains
49 fixed books, 725 reset windows and 1,475,654 targets. Neither the original
reserved test nor the new fairness holdout was scored for the added cells.
Table~\ref{tab:localsource} reports all ten endpoints. These NLLs use BF16
projections with FP32/TF32x3 attention; the cost study below separately rescores
its six history-model checkpoints in FP32 IEEE.

\begin{table}[!htbp]
\centering\small
\begin{tabular}{@{}llrrrr@{}}
\toprule
& & \multicolumn{2}{c}{Development NLL} & \multicolumn{2}{c}{Fixed-book NLL} \\
Local design & Local entries & Seed A & Seed B & Seed A & Seed B \\
\midrule
Per-layer & Current, repeated & 2.8353 & 2.8341 & 3.5044 & 3.5267 \\
Per-layer & 128-position history & 2.8257 & 2.8246 & 3.4858 & 3.5069 \\
\midrule
Paired input & Current, repeated & 2.8349 & 2.8334 & 3.4994 & 3.5263 \\
Paired input & 128-position history & 2.8258 & 2.8250 & 3.4957 & 3.5005 \\
\midrule
Shared KV & 128-position history & 2.8275 & 2.8301 & 3.4823 & 3.5133 \\
\bottomrule
\end{tabular}

\caption{Source-by-history endpoints and the additional shared-KV variant at
2.5B tokens. Lower NLL is better. The four cells reuse six earlier endpoints
and add two paired-input repeated-current endpoints. Development has 1,809 documents and
2,095,344 valid targets. Seeds A/B are fixed across rows.}
\label{tab:localsource}
\end{table}

\paragraph{Content benefit and source interaction.}
Define $H_s=L(s,\text{repeated current})-L(s,\text{history})$ and
$I=H_{\mathrm{paired}}-H_{\mathrm{per\text{-}layer}}$.
Here $H>0$ favors history. Figure~\ref{fig:localsource} plots the loss difference
$\Delta_s=-H_s$, so its interaction is $\Delta_{\mathrm{paired}}-
\Delta_{\mathrm{per\text{-}layer}}=-I$; the raw gain convention below is retained
for reproducibility.
On development text, mean history gains are $\HistoryDevIndependentGain$
and $\HistoryDevPairedGain$ nats for per-layer and paired inputs
(\HistoryDevIndependentPpl\% and \HistoryDevPairedPpl\% geometric-mean
perplexity reductions). Pairing slightly reduces the gain on both seeds:
$I=-0.000529$ and $-0.001086$ nats. On fixed books, both seeds retain positive
history gains, but interaction changes sign ($-0.014856$, $+0.005971$ nats).
The paired-input perplexity reductions are 0.37\% and 2.55\%.
These observations support retaining history under both input-source choices;
two seeds do not establish source equivalence or the absence of interaction.

For the history model alone, shared input changes mean development loss by
$\SourceInputDevDelta$ nats (\SourceInputDevPpl\% higher perplexity).
Both seeds fall within the prespecified descriptive margin $\pm0.005$ nats.
Shared local KV changes it by $\SourceKVDevDelta$ nats
(\SourceKVDevPpl\%): one seed lies inside that margin and one above it.
The prespecified rule requiring both seeds at least $+0.005$ nats for an
independent-source advantage is not met. This is an engineering margin, not
an equivalence test. Fixed-book changes against the history model have opposite directions
across seeds for both source variants.

\paragraph{Execution.}
Source models use BF16 projections, FP32/TF32x3 attention, tiled backward,
compiled loss and fused AdamW. Qualified rank changes preserve global batch,
data order and schedule. The original four endpoints use 58.64 GPU hours plus
0.03 for books; the two added controls use 28.29 plus 0.013, excluding separate
kernel and restoration qualification. Tasks and costs use the six history endpoints.

\paragraph{Supplemental task transfer.}
The three tasks and scoring rules are inherited from the protocol in
Appendix~\ref{app:taskprotocol}; all six checkpoints are evaluated on each task.
FP32 packing/batch checks cost 0.04 GPU hours and formal scoring 0.10.
The history model's per-example outputs match the earlier evaluation exactly.
Shared input changes two-seed mean accuracy by $+0.24$, $+0.24$ and
$+0.72$ percentage points on PIQA, HellaSwag and ARC-Easy; shared local KV
changes it by $-0.22$, $+0.29$ and $+0.59$ points. Both seeds favor both
variants on HellaSwag and ARC-Easy, while PIQA changes direction across seeds.
Every conditional example interval includes zero. The results characterize
these checkpoints' task differences; they establish neither a general task
advantage nor equivalence. Task contamination has not been audited.

\begin{table}[!htbp]
\centering\setlength{\tabcolsep}{4pt}
\begin{tabular}{@{}llrrr@{}}
\toprule
Local source & Task & A & B & Mean\;[95\% CI] \\
\midrule
Shared input & PIQA & $+0.65$ & $-0.16$ & $+0.24\;[-0.68, +1.17]$ \\
 & HellaSwag & $+0.35$ & $+0.14$ & $+0.24\;[-0.10, +0.59]$ \\
 & ARC-Easy & $+0.55$ & $+0.88$ & $+0.72\;[-0.25, +1.66]$ \\
\addlinespace[3pt]
Shared local KV & PIQA & $+0.05$ & $-0.49$ & $-0.22\;[-1.25, +0.79]$ \\
 & HellaSwag & $+0.53$ & $+0.05$ & $+0.29\;[-0.07, +0.64]$ \\
 & ARC-Easy & $+0.46$ & $+0.72$ & $+0.59\;[-0.44, +1.64]$ \\
\bottomrule
\end{tabular}

\caption{Source variant minus the history model at the same seed, in accuracy percentage points.
A/B are training-seed differences. Conditional paired-example intervals
follow Appendix~\ref{app:taskprotocol}.}
\label{tab:sourcetaskcontrasts}
\end{table}

\FloatBarrier
\subsubsection{Complete cache and exact execution with shared sources}
\label{app:sourcecost}
Source sharing changes which upper positions suffice for exact construction.
Pair $U$ upper layers into $G=U/2$ adjacent groups, indexed $g=1,\ldots,G$. Every receiver
in a group reads local entries projected from that group's entry representation.
An output suffix of length $r$ therefore requires at most $r+W-1$ entry
positions for the entire group: within-group query and residual updates
are positionwise once those banks and global KV are fixed. Applying this
recurrence backward from the final token gives
\begin{align}
 r_{\rm in}^{\rm group}(g)&=\min\{N,1+(G-g+1)(W-1)\},\\
 r_{\rm out}^{\rm group}(g)&=\min\{N,1+(G-g)(W-1)\}.
\end{align}
Every local bank retains its last $\min(N,W)$ entries, including the final
group. Shared input keeps separate receiver projections; shared local KV
projects, appends and stores each group's bank once. Lower-layer and global
KV remain complete. The schedule changes execution, not the trained model. For eight upper layers,
$W=128$ and $N=1{,}792$, it reduces the upper query/feedforward position count
from 3,564 to 1,532. The lower stack still processes the full prompt, so this
count reduction is distinct from end-to-end speedup.

\paragraph{Matched implementation and numerical checks.}
All six models use FP32 IEEE projections and attention, CUDA-graph prefill,
and eager exact continuation. Both full and exact routes are retained;
architectural comparisons always use the exact route for each model.
The grid uses batches 1/4/8 and prompts 512/1,024/1,792, followed by 128
teacher-forced decode steps (Appendix~\ref{app:costprotocol}).
A 54-condition CPU check covers packed boundaries, grouped sources, chunked
continuation, saved-state reload and physical tensor storage. Each trained
endpoint then passes 91 GPU checks: its original forward implementation,
full-sequence versus 128-step continuation, complete full/exact caches,
graph replay and output ownership, state restoration and physical storage.
The existing $10^{-4}$ absolute/relative logit and $10^{-6}$ average-NLL
tolerances are unchanged; graph comparisons are bitwise.

Each seed's three architectures run on one H20, with model order reversed
between seeds. Each route receives two warmups and ten timed repetitions;
full/exact order alternates. The study retains 1,080 timings across 54
model--workload cells. This controls device and route order while leaving
possible temporal drift. Repetitions measure execution variation, not
additional training seeds.

\paragraph{Quality and resource allocation.}
Table~\ref{tab:sourcecost} joins complete development NLL in the measured
precision with physical cache and request costs. At batch one and $N=1{,}792$,
sharing local KV reduces retained KV from 33.5 to 32.5 MiB; both add the same
document metadata. Thus halving the upper local banks removes about 3\% of
the complete retained state. Shared input retains the original bank count.
At prompt lengths 512/1,024/1,792, KV-only totals are 11/20/33.5 MiB for
the history model and shared-input variant versus 10/19/32.5 MiB for shared local KV; totals scale
with batch size $B$, with an additional $8BN$ bytes of document metadata.
At this workload, exact-prefill time ratios to the history model are
\ClosureInputPrefillRatio{} and \ClosureKVPrefillRatio{} for shared input
and shared local KV; complete-request ratios are
\ClosureInputRequestRatio{} and \ClosureKVRequestRatio{}.
Across the nine conditions, geometric-mean request ratios range from
\ClosureInputRequestMin{} to \ClosureInputRequestMax{} for shared input and
\ClosureKVRequestMin{} to \ClosureKVRequestMax{} for shared local KV.
The artifact retains both seeds throughout the grid, including all absolute times.
These measurements complement the source-quality contrasts; they do not
constitute a new timing comparison with GQA. FP32 fixed-batch reference
execution leaves BF16 and continuous-serving behavior open.

\begin{table}[!htbp]
\centering\small\setlength{\tabcolsep}{3pt}
\begin{tabular}{llrrrrrr}
\toprule
Source & Seed & FP32 NLL & Cache MiB & Prefill ms & Decode ms & Request ms & Peak$^*$ MiB \\
\midrule
Local history & A & 2.8256 & 33.514 & 24.94 & 1333.13 & 1358.13 & 53.72 \\
Local history & B & 2.8245 & 33.514 & 24.94 & 1345.41 & 1370.36 & 53.72 \\
Input & A & 2.8257 & 33.514 & 21.50 & 1342.32 & 1363.81 & 54.19 \\
Input & B & 2.8249 & 33.514 & 21.26 & 1341.05 & 1362.34 & 54.19 \\
KV & A & 2.8274 & 32.514 & 21.28 & 1184.32 & 1205.57 & 52.90 \\
KV & B & 2.8300 & 32.514 & 21.26 & 1201.08 & 1222.32 & 52.90 \\
\bottomrule
\end{tabular}

\caption{Local-source operating points at batch one, 1,792 prompt tokens and
128 decode tokens. Input/KV denote shared input/shared local KV. All models
use their exact schedule. Cache includes all lower/global/local KV and
metadata. $^*$Peak is incremental allocated memory above resident weights
and graph buffers, not total process memory.}
\label{tab:sourcecost}
\end{table}

\FloatBarrier

\subsection{Standard GQA alternatives at matched training computation}
\label{app:gqa}
This study compares six frozen endpoints: the history model, GQA2 and GQA4,
each at common-budget seeds A/B. The two history-model endpoints are the existing
2.5B-token anchors; four new GQA endpoints are trained from scratch. GQA4
uses four KV heads in every layer and 125,854,464 parameters. GQA2 uses
two KV heads and FFN width 2,144, giving 126,247,680 parameters compared
with the history model's 126,248,448. All models have width 768, 16 layers, 12 query
heads, head dimension 64, vocabulary 32,768 and context length 2,048.

\paragraph{Resources and common recipe.}
The data manifest, tokenizer, per-seed data-order prefix, effective batch of
64 windows, AdamW settings, warmup of 128 updates and peak learning rate
$3\times10^{-4}$ are shared. Each cosine schedule spans its own fixed budget.
GQA2 trains for 19,269 updates (2,525,626,368 input tokens); GQA4 trains for
19,323 (2,532,704,256). The first complete update meeting the history model's matrix-product
budget determines each endpoint, with less than 0.003\% excess. The count uses
actual causal, same-document attention pairs and forward/backward matrix
products; normalization, softmax, optimizer operations, padding and kernel
recomputation are outside this count. Thus it matches counted computation,
not elapsed time. The common recipe is not an independent hyperparameter
optimum for each architecture.

\paragraph{Qualification and scoring.}
Full-shape kernel/gradient checks and bitwise 256 versus fresh-process 128+128
restoration qualify each new model. Independent endpoint audits verify source
and checkpoint identities, contiguous updates, schedules and batch counts.
Scoring uses the development and 49-book conditions from Appendix~\ref{app:data};
the reserved test is unchanged. Book comparisons bind checkpoint identities
before scoring.

\begin{table}[!htbp]
\centering
\begin{tabular}{llrrr}
\toprule
Control & Seed & Dev. & Books & Book 95\% CI \\
\midrule
GQA2 & A & $-24.9$ & $-77.9$ & $[-89.3,-68.6]$ \\
 & B & $-18.1$ & $-37.9$ & $[-42.9,-33.3]$ \\
\addlinespace[3pt]
GQA4 & A & $-8.6$ & $-39.0$ & $[-47.1,-32.2]$ \\
 & B & $-10.0$ & $+0.2$ & $[-5.2,+6.0]$ \\
\bottomrule
\end{tabular}

\caption{NLL differences in $10^{-3}$ nats. The history model minus each standard GQA alternative. Negative differences favor
the history model. Book intervals resample 49 books with 2,000 paired draws (seed 777),
conditional on the fixed checkpoints; they are not training-seed intervals.}
\label{tab:gqacontrasts}
\end{table}

Both seeds favor the history model on development text and against GQA2 on the fixed
books. Against GQA4, seed A favors the history model and seed B is nearly tied with a
conditional book interval spanning zero. Means across two seeds summarize
these endpoints; the documents and windows do not increase the independent
training sample size. The four new runs allocate 53.85 GPU hours to training
and endpoint validation; external evaluation adds 0.02 GPU hours.

\paragraph{Downstream tasks.}
All six endpoints use the fixed FP32 scoring protocol in
Appendix~\ref{app:taskprotocol}. Packing and batch-invariance checks pass the
$10^{-3}$ token-log-probability tolerance on training fixtures; reused history-model
per-example outputs reproduce the earlier records bitwise. Table~\ref{tab:gqataskcontrasts} reports all paired task contrasts;
absolute and secondary-metric scores accompany the artifact.

\begin{table}[!htbp]
\centering
\begin{tabular}{@{}llrrr@{}}
\toprule
Task & Control & A & B & Mean\;[95\% CI] \\
\midrule
PIQA & GQA2 & $-0.05$ & $+0.33$ & $+0.14\;[-0.95,+1.20]$ \\
 & GQA4 & $-0.05$ & $+1.36$ & $+0.65\;[-0.38,+1.66]$ \\
\addlinespace[3pt]
HellaSwag & GQA2 & $+0.04$ & $-0.04$ & $+0.00\;[-0.38,+0.38]$ \\
 & GQA4 & $+0.12$ & $+0.23$ & $+0.17\;[-0.21,+0.57]$ \\
\addlinespace[3pt]
ARC-Easy & GQA2 & $-1.85$ & $-1.52$ & $-1.68\;[-2.71,-0.65]$ \\
 & GQA4 & $-1.98$ & $-0.84$ & $-1.41\;[-2.44,-0.38]$ \\
\bottomrule
\end{tabular}

\caption{The history model minus GQA accuracy in percentage points. Paired-example
intervals condition on these checkpoints (Appendix~\ref{app:taskprotocol});
A/B preserve both training-seed differences.}
\label{tab:gqataskcontrasts}
\end{table}

PIQA and HellaSwag contrasts have conditional intervals spanning zero.
ARC-Easy favors GQA2 and GQA4 by 1.68 and 1.41 percentage points on average,
with the same direction in both seeds and conditional intervals excluding
zero. The likelihood advantage therefore does not establish a general
zero-shot task advantage. Public benchmark overlap with training data remains
unaudited; these are supplemental architecture comparisons.

\FloatBarrier
\subsection{Common-recipe adjacent-layer sharing}
\label{app:clacommon}
This study compares two ways of sharing KV with the shared-global/local
allocation. GQA2 shares within a layer's query heads; \CLABaseline{} shares a projected
bank across each adjacent pair of layers. In \CLABaseline{}, the first layer forms KV
from its own normalized input, and the second reads the same bank with its own
query. Sixteen layers therefore retain eight full-length banks with four KV
heads each. GQA2 retains sixteen banks with two heads each. Both have 126,247,680
parameters, using FFN width 2,144, hidden width 768 and twelve query heads.
We use the original naming convention: GQA4 denotes four KV heads and CLA2
a sharing factor of two. The artifact alias \texttt{CLA4} maps to this same
configuration. This common-recipe comparison does not search each family's
best structure.

\paragraph{Training budget and quality.}
\CLABaseline{} uses the same peak learning rate $3\times10^{-4}$, 128-step warmup,
cosine schedule, global batch 64 and 2K packing as the GQA comparison.
Seeds A/B follow the common-budget pairing. Each \CLABaseline{} endpoint takes 19,269 steps
and 2,525,626,368 input tokens; the counted training matrix-product budget
matches the history model to within 0.0023\%. Token counts and wall-clock training costs
are not matched. Both checkpoints pass a fresh-process, full-state 256-step
continuous versus 128+128 restoration check. Training and these restoration
checks use 26.57 allocated GPU hours.

Independent reload precedes scoring the original development text and fixed
49-book condition (725 windows, 1,475,654 targets). No additional LM test
split is opened. Table~\ref{tab:claendpoints} retains every endpoint. The history model's
mean paired NLL differences against \CLABaseline{} are $-0.02170$ on development text
and $-0.04339$ on books, with both seeds favoring the history model. These are descriptive
two-seed comparisons; the eight-seed history minus current-only book interval answers a different
contrast and remains unchanged. \CLABaseline{} versus GQA2 changes direction across seeds
on both text conditions.

\begin{table}[!htbp]
\centering\small
\begin{tabular}{@{}llrrrrr@{}}
\toprule
Design & Seed & Dev. NLL & Books NLL & PIQA & HellaSwag & ARC-E \\
\midrule
Local history & A & 2.8257 & 3.4858 & 61.32 & 30.58 & 42.47 \\
Local history & B & 2.8246 & 3.5069 & 62.19 & 30.77 & 42.85 \\
GQA2 & A & 2.8506 & 3.5637 & 61.37 & 30.54 & 44.32 \\
GQA2 & B & 2.8427 & 3.5448 & 61.86 & 30.81 & 44.36 \\
GQA4--CLA2 & A & 2.8455 & 3.5313 & 60.17 & 30.04 & 42.93 \\
GQA4--CLA2 & B & 2.8483 & 3.5481 & 60.94 & 30.56 & 43.22 \\
\bottomrule
\end{tabular}

\caption{All six endpoints under the common recipe. Language NLL uses the
training/scoring precision (BF16 projections with FP32/TF32x3 attention).
Task entries are accuracy percentages: raw PIQA, character-normalized
HellaSwag and ARC-Easy. A/B denote paired training seeds.}
\label{tab:claendpoints}
\end{table}

\paragraph{Fixed downstream tasks.}
The same six checkpoints receive PIQA, HellaSwag and ARC-Easy evaluation using
the frozen versions, templates and scoring rules in Appendix~\ref{app:taskprotocol}.
All twelve history/GQA2 endpoint--task records reproduce the earlier per-example
scores bitwise. Packed/unpacked and batch-invariance checks precede scoring.
Relative to \CLABaseline{}, PIQA improves by $1.20$ percentage points
(conditional 95\% interval $[0.08,2.31]$). HellaSwag changes by $+0.37$
points ($[0.00,0.75]$) and ARC-Easy by $-0.42$ ($[-1.49,0.61]$).
The latter two intervals include zero. The shared conditional protocol is
in Appendix~\ref{app:taskprotocol}; training-data contamination remains unaudited.

Complete-cache and request measurements for these checkpoints are in
Appendix~\ref{app:clacost}.

\FloatBarrier
\subsection{Task transfer of the budget controls}
\label{app:downstream}
We evaluate the eight completed budget endpoints in
Appendix~\ref{app:budgetprotocol}, with no weight updates or checkpoint
selection. The shared task protocol is in Appendix~\ref{app:taskprotocol};
these are the common-budget seeds A/B. The tasks and comparisons were specified after
the language-modeling study, before any downstream model scores were read;
this is a supplemental assessment, not a new preregistered confirmation of
the original held-out study.

\paragraph{Paired differences and uncertainty.}
Table~\ref{tab:downstreamcontrasts} retains both seed differences; positive
values favor the history model. Intervals follow the conditional paired-example protocol in
Appendix~\ref{app:taskprotocol}. Every history-versus-compute-matched interval includes zero, leaving the task-accuracy
contrast unresolved.

\begin{table}[!htbp]
\centering
\begin{tabular}{@{}llrrr@{}}
\toprule
Task & Control & A & B & Mean\;[95\% CI] \\
\midrule
PIQA & Current + updates & $-0.27$ & $+1.36$ & $+0.54\;[-0.46,+1.58]$ \\
 & Current only & $+0.60$ & $+1.25$ & $+0.92\;[-0.16,+1.99]$ \\
 & Repeated current & $+0.44$ & $+1.85$ & $+1.14\;[+0.11,+2.20]$ \\
\addlinespace[3pt]
HellaSwag & Current + updates & $+0.03$ & $-0.10$ & $-0.03\;[-0.41,+0.33]$ \\
 & Current only & $-0.18$ & $+0.25$ & $+0.03\;[-0.33,+0.40]$ \\
 & Repeated current & $-0.42$ & $-0.35$ & $-0.38\;[-0.772,-0.005]$ \\
\addlinespace[3pt]
ARC-Easy & Current + updates & $-1.22$ & $-0.21$ & $-0.72\;[-1.75,+0.34]$ \\
 & Current only & $-0.55$ & $-0.42$ & $-0.48\;[-1.52,+0.53]$ \\
 & Repeated current & $-1.35$ & $+0.17$ & $-0.59\;[-1.58,+0.38]$ \\
\bottomrule
\end{tabular}

\caption{The history model minus each control in accuracy percentage points, using each
task's fixed primary metric. The last column quantifies conditional example
variation, not training-seed variation. All comparisons are shown.}
\label{tab:downstreamcontrasts}
\end{table}

\FloatBarrier
\section{Execution and complete-state costs}
\label{app:resources}
Each comparison combines quality and resources from the same checkpoints.
Selected-rate results are in Appendix~\ref{app:finallr}, and source-sharing
results in Appendix~\ref{app:sourcecost}. This section gives the complementary
common-recipe comparisons, followed by exact construction and state retention.
All use Appendix~\ref{app:costprotocol}; timings stay within their own campaign.

\FloatBarrier
\subsection{Common-recipe adjacent-layer sharing costs}
\label{app:clacost}
The six checkpoints are those in Appendix~\ref{app:clacommon}.

\paragraph{Matched execution.}
All six checkpoints use the common FP32 protocol
(Appendix~\ref{app:costprotocol}); the four reused history/GQA2 scores reproduce
their earlier FP32 values exactly. Trained-forward agreement, cache equivalence,
continuation and restoration are qualified before timing. CLA stores each
producer bank once. Its final pair computes
full producer KV but only the last query, output and FFN position; both full
and reduced-query routes pass cache and continuation checks.

Three blocks rotate architecture order. Each uses two warmups and four
alternating full/exact repeats, over the nine 128-step cells and 1/8/32-step
extensions at $(B,N)=(1,1792)$ and $(4,1024)$. The 2,160 timings use
1.10 worker GPU hours. Times are geometric means of block medians and seeds;
NLL is an arithmetic mean. Table~\ref{tab:claprimary} uses this earlier campaign; its fifteen-cell
request grid accompanies the artifact. Figure~\ref{fig:gqatradeoff} instead uses the selected-rate
checkpoints and independent timing campaign in Appendix~\ref{app:finallr}.

\begin{table}[!htbp]
\centering\small
\begin{tabular}{@{}lrrrr@{}}
\toprule
Design & FP32 dev. NLL & Cache (MiB) & Prefill (ms) & Request (s) \\
\midrule
Local history & 2.8251 & 33.51 & 24.93 & 1.329 \\
GQA2 & 2.8465 & 28.01 & 32.11 & 1.163 \\
GQA4--CLA2 & 2.8468 & 28.01 & 29.86 & 1.009 \\
\bottomrule
\end{tabular}

\caption{Same-checkpoint FP32 quality and cost at batch one, 1,792 prompt
positions and 128 extra decode steps. Cache is complete post-prefill storage,
not total device memory. This campaign is separate from the five-design one.}
\label{tab:claprimary}
\end{table}

Across the nine 128-step cells, history/\CLABaseline{} request-time ratios range from
1.227 to 1.330. Short-request ordering depends on batch and prompt shape:
at batch one with 1,792 prompt positions the history model is faster with one extra step,
whereas \CLABaseline{} is already faster with one at batch four and 1,024 positions.
The comparison identifies a predictive-quality premium with measurable storage
and execution costs, rather than a common speed advantage across workloads.

\FloatBarrier
\subsection{Unified five-design quality--cost comparison}
\label{app:unifiedcost}
This complementary comparison remeasures all ten checkpoints from the history model, shared-input, shared-local-KV,
GQA2 and GQA4 configurations in one campaign. Each seed uses one H20; model order
rotates over three blocks and the initial order reverses across seeds. All
models pass qualification before any timing block begins. Historical timings
are not combined with this campaign. The earlier focused studies below retain
their original numerical records.

\paragraph{Measurement protocol.}
The ten checkpoints follow Appendix~\ref{app:costprotocol}, with full/exact
cache, continuation, graph ownership and restoration qualification.
Element/logit tolerances are $10^{-4}$
absolute/relative, mean NLL tolerance is $10^{-6}$, and graph checks are bitwise.
Physical cache storage agrees with tensor counts.

The nine 128-step cells are extended by 1/8/32-step measurements at
$(B,N)=(1,1792)$ and $(4,1024)$. Three rotated blocks each have two warmups and
four alternating full/exact repeats: 3,600 timings and 1.92 worker GPU hours,
excluding qualification and orchestration. Ratios aggregate block medians
geometrically across blocks and paired seeds.

\begin{figure}[!ht]
\centering
\includegraphics[width=\linewidth]{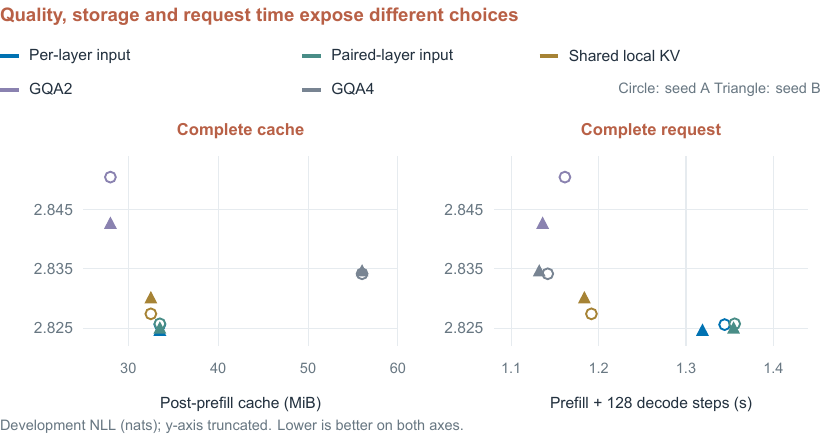}
\caption{\textbf{Input sharing, bank sharing and GQA offer different quality--cost choices.}
Circles/triangles denote two seeds at 126M scale and 2K training context;
lower is better on both axes. Each
checkpoint supplies FP32 development NLL, complete cache and request time.
Cache includes lower/global/local KV and document IDs. Requests use batch one,
a 1,792-token prompt and 128 extra teacher-forced decode steps, with graph
prefill and eager decode. Times aggregate three rotated blocks.}
\label{fig:fivefamilycost}
\end{figure}

\begin{table}[!htbp]
\centering\small
\begin{tabular}{@{}lrrrr@{}}
\toprule
Design & Dev. NLL & Cache (MiB) & Prefill (ms) & Request (s) \\
\midrule
Local history (per-layer) & 2.8251 & 33.51 & 24.96 & 1.331 \\
Paired input & 2.8253 & 33.51 & 21.49 & 1.355 \\
Shared local KV & 2.8287 & 32.51 & 21.33 & 1.188 \\
GQA2 & 2.8465 & 28.01 & 32.13 & 1.149 \\
GQA4 & 2.8344 & 56.01 & 33.29 & 1.137 \\
\bottomrule
\end{tabular}

\caption{Unified campaign at batch one, prompt 1,792 and 128 extra decode
steps. NLL is the two-seed arithmetic mean; times use geometric aggregation
of block medians and seeds. Cache is complete post-prefill tensor storage.
Figure~\ref{fig:fivefamilycost} retains individual seed points.}
\label{tab:unifiedprimary}
\end{table}

Across the nine 128-step cells, history/GQA2 request ratios range from
\UnifiedTwoRequestMin{} to \UnifiedTwoRequestMax{}, and history/GQA4 from
\UnifiedFourRequestMin{} to \UnifiedFourRequestMax{}. At the two short-output
shapes, the one-step and eight-step measurements reverse the GQA/history timing
order in the two-seed mean. GQA4 at batch one is nearly tied for seed B at
eight steps (GQA4/history $=1.002$). This brackets a workload-dependent transition;
it does not locate a
universal crossover or establish continuous-batching performance. The observed
quality, storage and time choices apply to these fixed-batch FP32 implementations.

\FloatBarrier
\paragraph{Earlier fixed-order GQA measurement.}
\label{app:gqacost}
An earlier six-endpoint history/GQA2/GQA4 campaign used the same nine 128-step
workloads with two warmups and ten alternating route repeats (1,080 timings).
Each seed used one H20, with a fixed architecture order. Across the grid,
history/GQA2 request-time ratios were 1.083--1.171 and history/GQA4 ratios were
1.054--1.153. The history model used more cache than GQA2 and less than GQA4.
The later rotated campaigns above provide the current architecture-cost
comparisons. The companion's \texttt{historical-gqa.html} retains the full
earlier protocol and all six endpoint, nine aggregate and eighteen seed-level
rows, including graph setup and memory-accounting definitions. These records
are reported separately because the ordering and repetition schemes differ.

\FloatBarrier
\subsection{Quality and request cost at the same checkpoints}
\label{app:matchedcost}
\begin{figure}[!ht]
\centering
\includegraphics[width=\linewidth]{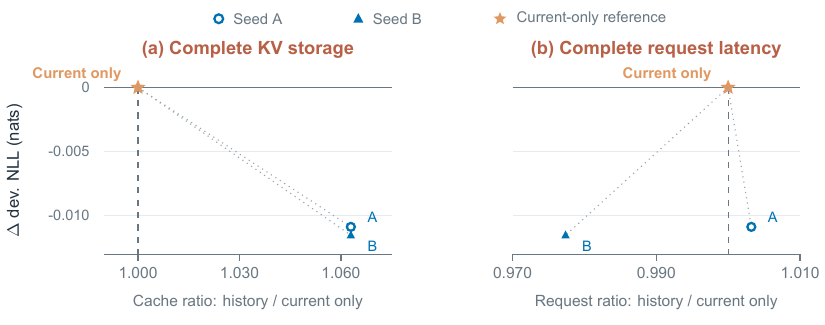}
\caption{\textbf{Adding history and accelerating its construction are different
comparisons.} The same four 2.5B-token checkpoints supply development quality
and cost. Panels (a--b) compare exact history and current-only execution at batch one,
1,792 prompt tokens and 128 teacher-forced decode steps. Each history-model point is a
paired seed; the current-only control is the reference at zero NLL difference and unit cost. Dotted
segments connect comparisons, not intermediate models; lower values on both
axes are better. Separately, exact execution within the history model gives $1.82\times$
prefill speed and $1.014\times$ complete-request speed relative to full execution
(geometric means across seeds). All costs use matched FP32 execution; the full nine-condition
grid appears in Appendix~\ref{app:matchedcost}.}
\label{fig:qualitycost}
\end{figure}

The original system measurements and factorial quality results use different
training recipes. Here, four existing 2.5B-token checkpoints (the history model and the current-only control,
seeds A/B) supply both quality and cost measurements in the same FP32 IEEE
implementation. A/B follow the common-budget pairing. A trained-model
logit/loss bridge qualifies the scoring implementation before measurement.
Reaggregating all development targets gives the losses in
Figure~\ref{fig:qualitycost}: the history model lowers geometric-mean perplexity by
1.11\%. These are additional measurements of existing checkpoints, not new
training seeds.

\paragraph{Request measurements.}
This four-checkpoint campaign follows Appendix~\ref{app:costprotocol}, using
the nine 128-step cells. Before timing, 220 checks cover logits, full cache,
continuation, document isolation and graph reuse. Two warmups and ten
interleaved full/exact repeats give 720 timings. Graph setup time and storage
are recorded separately.

Let $S_p$ and $S_r$ denote full/exact latency within the same history model,
for prefill and the complete request, respectively. Let $R_p$ and $R_r$
denote history/current-only latency under exact execution for prefill and the request; $R_c$ is
their complete post-prefill KV byte ratio. Larger $S$ means a faster exact
schedule; larger $R$ means greater cost for the history model. Each ratio uses the median
of ten timings, then Table~\ref{tab:matchedgrid} takes its geometric mean
over the two checkpoints. Full and exact routes retain identical complete
caches for each model.

\begin{table}[!htbp]
\centering\small
\begin{tabular}{rrrrrrr}
\toprule
Batch & Prompt & $S_p$ & $S_r$ & $R_p$ & $R_r$ & $R_c$ \\
\midrule
1 & 512 & 1.137 & 1.000 & 1.492 & 1.005 & 1.220 \\
1 & 1024 & 1.391 & 1.002 & 1.474 & 0.995 & 1.110 \\
1 & 1792 & 1.822 & 1.014 & 1.311 & 0.990 & 1.063 \\
4 & 512 & 1.163 & 1.002 & 1.732 & 0.993 & 1.220 \\
4 & 1024 & 1.440 & 1.014 & 1.535 & 0.988 & 1.110 \\
4 & 1792 & 1.935 & 1.058 & 1.351 & 1.016 & 1.063 \\
8 & 512 & 1.185 & 1.009 & 1.736 & 1.008 & 1.220 \\
8 & 1024 & 1.460 & 1.029 & 1.550 & 1.006 & 1.110 \\
8 & 1792 & 1.924 & 1.105 & 1.325 & 1.056 & 1.063 \\
\bottomrule
\end{tabular}

\caption{All nine request conditions. The $S$ columns compare execution
schedules within the history model; the $R$ columns compare the history model with the current-only control. Cache counts
include lower, shared global and upper local KV.}
\label{tab:matchedgrid}
\end{table}

At the reference workload, history/current-only request ratios differ by seed
(A: 1.003; B: 0.977). The within-model prefill
improvement is therefore distinct from an architecture-level request-speed
advantage. Decode time determines how much of a prefill saving remains in
the complete request.

The artifact retains both checkpoints at all nine shapes and the full published
ratios. Timing repetitions describe execution variation within a checkpoint,
not additional independent training seeds.

\FloatBarrier
\subsection{Exact construction and state retention}
\label{app:executionfigures}

\begin{figure}[!htbp]
\centering
\includegraphics[width=\linewidth]{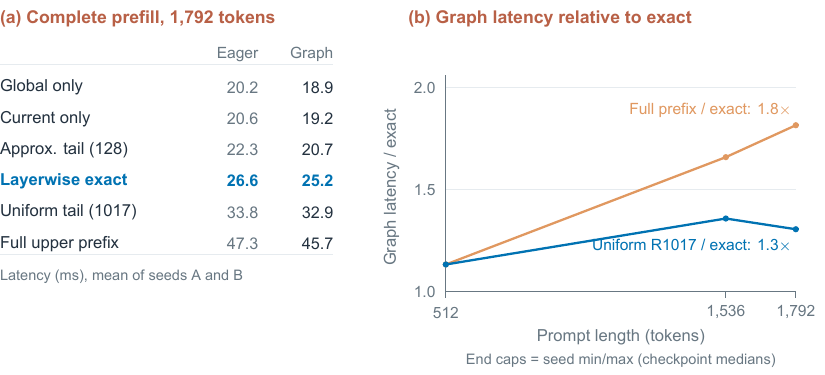}
\caption{\textbf{Exact construction reduces the history model's prefill cost.}
(a) Columns give mean latency in ms across two checkpoint medians at 1,792
tokens, under Eager and CUDA Graph execution. The blue row is layerwise exact
execution. All six routes receive
matched acceleration. (b) Graphed latency relative to layerwise exact, preserving
the same complete history-model cache; points and lines show two-checkpoint means, with
small caps spanning the checkpoint range (not confidence intervals). These H20
measurements use FP32 attention, batch one and 15 repeats per median, including
device-input and owned-output copies, excluding loading and graph creation.}
\label{fig:graph}
\end{figure}

\begin{figure}[!htbp]
\centering
\includegraphics[width=\linewidth]{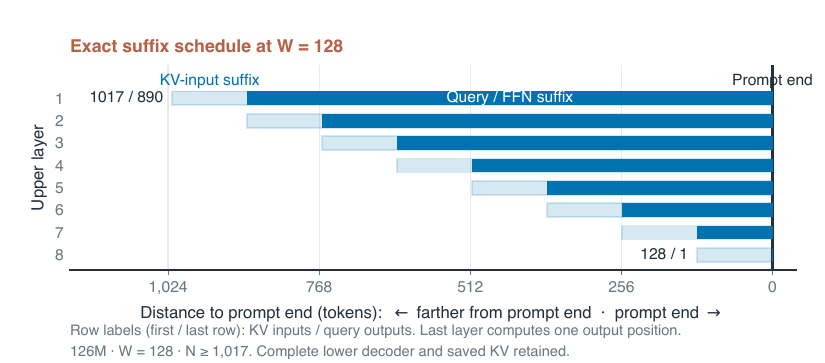}
\caption{The sufficient 126M schedule distinguishes KV inputs from query/FFN
outputs at each upper layer. The right-aligned bands share the same prompt end. All
lower/global and saved local KV are retained; the counts are not measured speedups.}
\label{fig:schedule}
\end{figure}
\begin{table}[!htbp]
\centering
\begin{tabular}{lrr}
\toprule
Prefill route & A (ms) & B (ms) \\
\midrule
Global only, last position & 18.880 & 18.961 \\
Current only, last position & 19.142 & 19.195 \\
Local history, approximate $R=128$ & 20.660 & 20.727 \\
Local history, layerwise exact & 25.141 & 25.247 \\
Local history, uniform sufficient $R=1017$ & 32.857 & 32.937 \\
Local history, full upper prefix & 45.665 & 45.798 \\
\bottomrule
\end{tabular}

\caption{End-to-end cold-prefill medians at $N=1792$, with CUDA Graphs on every
route. Inputs and weights are device-resident, batch size is one, and attention
uses explicit FP32 operations. The $R128$ route is approximate; the other history-model
routes preserve the same complete cache within the numerical qualification.}
\label{tab:prefill}
\end{table}

\subsubsection{Retention and reconstruction of local state}
\label{app:retention}
\paragraph{Complete KV footprint.}
Each lower decoder layer retains its full causal KV.
For $L$ lower and $U$ upper layers, batch size $B$, KV width $k$ for each of
K and V, and $b_k$ bytes per element, the logical KV tensor sizes are
\begin{equation}
 S_{\rm shared/local}=2Bk b_k\big[(L+1)N+U\min(N,W)\big],
 \qquad S_{\rm independent}=2Bk b_k(L+U)N.
 \label{eq:totalcache}
\end{equation}
The extra full-length term is the shared global provider, counted once.
At $L=U=8$, $k=256$, $W=128$, $N=1792$, $B=1$ and FP32, these are
\KVSharedMiB\ and \KVIndependentMiB\ MiB, a \KVReductionFactor-fold reduction.
Including the document IDs gives the measured 33.514 MiB in
Table~\ref{tab:lifecycle}. These counts exclude weights, temporary workspaces
and allocator overhead. As $N$ increases with this dense lower decoder fixed,
the KV ratio approaches $16/9\approx\KVAsymptoticFactor$: the lower decoder
and shared provider retain linear growth, while upper local storage is bounded.

\paragraph{Replacing local state with boundary representations.}
Boundary replay replaces upper local KV during a pause with $R$ lower-boundary
representations for later replay, in addition to all lower/global KV and document
metadata. For batch $B$, upper depth $U$, window $W$, KV width $k$ for each of
K and V, and bytes per boundary/KV element $b_h,b_k$, the replaced and added
tensor storage is
\begin{equation}
 S_{\rm local}=2BUWk b_k,\quad S_{\rm boundary}=BRd b_h,
 \qquad R<\frac{2UWk b_k}{d b_h}\quad\text{for positive savings.}
 \label{eq:storage}
\end{equation}
This assumes prompts of at least $W$ positions and uncompressed boundary states;
it excludes common state and transient workspaces. A sufficient replay suffix
for all saved local KV is $R_s=W+(U-1)(W-1)=1017$ in the tested backbone.
With $k=256,d=768$ and equal precision, positive savings instead require
$R<682.667$. Keeping the local KV therefore uses less tensor storage than saving the
boundary horizon for this sufficient exact reconstruction protocol.

Approximate $R128$ replay was selected on an eight-window screen and evaluated on
64 new development windows at two pause positions per window. It gives first-32
NLL changes of $-0.000066/-0.000120$ relative to retained exact state; dropping
local state without replay instead gave $+0.040701/+0.056706$. These near-zero changes show close reconstruction in the tested windows.
Including the added boundary states, total retained cache tensors decrease by
4.85\%--5.60\% over the tested prompt lengths.

\begin{table}[!htbp]
\centering
\begin{tabular}{lrrrr}
\toprule
Pause strategy & GPU MiB & Host MiB & Resume ms (A) & (B) \\
\midrule
Keep full cache & 33.514 & 0.000 & 10.880 & 10.839 \\
Offload upper local & 31.514 & 2.000 & 10.976 & 10.813 \\
Offload all cache & 0.000 & 33.514 & 11.232 & 11.167 \\
Boundary replay 128 & 31.889 & 0.000 & 16.353 & 16.202 \\
Recompute from tokens & 0.000 & 0.000 & 57.440 & 57.600 \\
\bottomrule
\end{tabular}

\caption{Reference pause/resume costs at 1,792 prompt tokens. Resume includes
restoration and the next cached prediction. GPU/host columns count retained cache
tensors; all routes also keep prompt tokens on the host. These FP32, immediate,
single-request pause measurements are separate from the graph-prefill study.}
\label{tab:lifecycle}
\end{table}

The lifecycle measurements compare full retention, local-only CPU offload,
full CPU offload, boundary replay and recomputation from tokens. Cache-only replay
omits the top layer's unnecessary output computation and passes CPU and H20
qualification. Table~\ref{tab:lifecycle} identifies the resource trade-off: $R128$
avoids 2 MiB of host cache at the cost of 0.375 MiB more paused GPU state and
1.49--1.50 times the local-offload restoration latency. Local CPU offload is
therefore the lower-latency, lower-GPU-storage option in this immediate,
single-request condition.

\begin{figure}[!htbp]
\centering
\includegraphics[width=\linewidth]{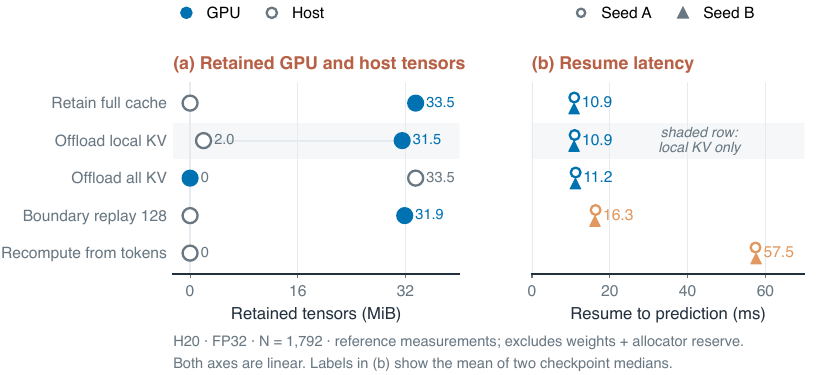}
\caption{\textbf{Retained state and reconstruction time measure different costs.}
The same reference pause/resume measurements as Table~\ref{tab:lifecycle},
at batch one and 1,792 prompt tokens. (a) Filled blue and open gray points show retained GPU
and host tensors, respectively, including complete lower/global KV wherever
retained. Labels give component sizes; the shaded row offloads only
local KV. Weights and allocator reserve are excluded. (b) Circles and triangles
show the two checkpoint medians on a linear time axis; labels give their mean. Local offload retains 2 MiB on
the host and has lower observed resume latency than boundary replay. These are
FP32 H20 reference costs, with no training-seed confidence intervals.}
\label{fig:lifecycle}
\end{figure}

\FloatBarrier
\section{Artifact coverage and experiment identities}
\label{app:artifacts}

\FloatBarrier
\label{app:provenance}
The evaluated dense decoder builds on an internal architecture report and is
specified fully here. The report's broader MoE and multimodal designs are
outside these experiments. The companion retains provenance identifiers.

The archive README gives commands, dependencies and a coverage map.
Validation records document bitwise restart checks for three architectures
and development-score checks for six original checkpoints. The archive
includes selected-rate raw scores and historical aggregates; weights,
source text and complete historical raw streams are excluded.

\paragraph{Complete records.}
The artifact's \texttt{complete\_tables/index.json} maps every original table to
its revised location or archived record. Expanded TeX and row/cell CSV retain
all published digits, captions and conditions. They preserve publication-level
records; historical raw scoring streams are included only where listed in the
coverage map. Current results and figures are bound to the manuscript hash.

\subsection{Training replicates and checkpoint reuse}
\label{app:seedindex}
Letters identify paired runs within a study; they are not global identifiers.
The factorial uses eight independent seeds. Common-budget comparisons reuse
compatible checkpoints, and selected-rate confirmation uses a new pair.
The 8K study continues those confirmation parents, adding no independent seed.
Table~\ref{tab:seedindex} summarizes this relationship. The artifact's
\texttt{seed\_identity.json} maps every letter to its original integer seed,
study family and endpoint records.
Bootstrap RNG seeds (777 or 916, where specified) control resampling, not training.

\begin{table}[H]
\centering\small
\begin{tabular}{@{}lll@{}}
\toprule
Study family & Labels & Relationship \\
\midrule
Content $\times$ fusion & A--H & Eight independent seeds \\
Budget / sources / common recipe & A/B & Compatible endpoints reused \\
Learning-rate selection & Common A & One shared selection seed \\
Selected-rate confirmation & A/B & Two new seeds \\
8K adaptation & Confirmation A/B & Continues the same parents \\
Earlier controls / 305M / provider & A/B & Earlier paired seeds \\
Initial discovery & Discovery & Separate from confirmation \\
\bottomrule
\end{tabular}
\caption{Training-replicate identities. Letters are local to a study family;
new evaluations of a checkpoint do not create new independent runs. Exact seeds
and checkpoint identifiers are retained in the artifact.}
\label{tab:seedindex}
\end{table}

\subsection{Intervention identifiers}
\label{app:controlnames}
The manuscript uses descriptive names; configuration keys remain unchanged.
The factorial uses $J_W$ and $S_W$ for joint and separate fusion at window $W$.
\begin{center}
\small
\begin{tabular}{@{}ll@{}}
\toprule
Configuration key & Manuscript name \\
\midrule
W128 / B2 & Local history \\
W1 & Current only \\
W1M & Repeated current \\
W1-C & Current + updates \\
B1 & Global only \\
B1C & Global + adapters \\
B0 & Independent KV \\
CLA4 & GQA4--CLA2 \\
\bottomrule
\end{tabular}
\end{center}

\end{document}